\documentclass[10pt,journal,compsoc]{IEEEtran}

\usepackage{microtype}
\usepackage{graphicx}
\usepackage{subcaption}
\usepackage{booktabs} 
\usepackage{bm,amsfonts,amsmath}
\usepackage{multirow}

\usepackage{hyperref}

\usepackage{algorithm}
\usepackage{algorithmic}
 \usepackage{soul} 

\usepackage{optidef}

\usepackage{array}
\newcolumntype{H}{>{\setbox0=\hbox\bgroup}c<{\egroup}@{}}
\usepackage{float}
\usepackage{color}
\usepackage{caption}
\usepackage{anyfontsize}
\usepackage{pgf}
\usepackage{enumitem}

\usepackage{amsmath}
\usepackage{amssymb}
\usepackage{mathtools}
\usepackage{amsthm}

\theoremstyle{plain}

\theoremstyle{definition}

\theoremstyle{remark}

\ifCLASSOPTIONcompsoc
  \usepackage[nocompress]{cite}
\else
  \usepackage{cite}
\fi

\ifCLASSINFOpdf

\else
 
\fi

\begin{document}

\title{Tensor-based Multi-view Spectral Clustering via Shared Latent Space }

\author{Qinghua Tao,~Francesco Tonin,~Panagiotis Patrinos, and~Johan~A.K. Suykens
\IEEEcompsocitemizethanks{\IEEEcompsocthanksitem Q. Tao, F. Tonin, P. Patrinos, J.A.K. Suykens are with STADIUS, ESAT, KU Leuven, Heverlee 3001, Belgium.
E-mail: \{qinghua.tao, francesco.tonin, panos.patrinos, johan.suykens\}@esat.kuleuven.be.\protect\\
Corresponding authors: Qinghua Tao and Francesco Tonin.
}
}

\markboth{}%
{Shell \MakeLowercase{\textit{et al.}}: Bare Advanced Demo of IEEEtran cls for IEEE Computer Society Journals}

\IEEEtitleabstractindextext{%
\begin{abstract}
Multi-view Spectral Clustering (MvSC) attracts increasing attention due to  diverse data sources. However, most existing works are prohibited in out-of-sample predictions and overlook model interpretability  and  exploration of  clustering results. In this paper, a new method for MvSC is proposed via a shared latent space from the Restricted Kernel Machine framework. Through the lens of conjugate feature duality, we cast the  weighted kernel principal component analysis problem for MvSC and develop a modified weighted conjugate feature duality to formulate  dual variables. In our method, the dual variables, playing the role of hidden features, are  shared by all views to construct a common latent space, coupling the  views by learning projections from  view-specific spaces. Such single latent space promotes well-separated clusters and provides straightforward data exploration, facilitating  visualization and interpretation. Our method requires only a single eigendecomposition, whose dimension is independent of the number of views. To boost higher-order correlations, tensor-based modelling is introduced  without increasing computational complexity. Our method can be flexibly applied with out-of-sample extensions, enabling greatly improved efficiency for large-scale data with fixed-size kernel schemes. Numerical experiments verify that our method is effective regarding accuracy, efficiency, and interpretability, showing a sharp eigenvalue decay and distinct latent variable distributions.
\end{abstract}

}

\maketitle

\IEEEdisplaynontitleabstractindextext

\IEEEpeerreviewmaketitle

\ifCLASSOPTIONcompsoc
\IEEEraisesectionheading{\section{Introduction}\label{sec:introduction}}
\else
\section{Introduction}
\label{sec:introduction}
\fi

\IEEEPARstart{T}{he} task of clustering aims to group data based on some similarity  {metrics}
between different data points, among which spectral clustering \cite{von2007tutorial}  has been attracting wide interest  due to its well-designed mathematical formulations and relations to graph theory \cite{chung1997spectral}.  Spectral clustering  commonly makes use of the eigenvectors of a normalized affinity matrix, e.g., the Laplacian matrix, to partition a corresponding graph based on the normalized minimum cut problem \cite{shi2000normalized}, where each data point corresponds to a node and the edge between paired nodes is regarded as their connectivity \cite{ng2002}.
In real-world applications, data can be collected from diverse sources and thereby described in different views.  {In this way},   multiple views are provided to represent complementary features of the data. For example, an image classification task with infrared and visible images  due to different measurement methods \cite{ma2019}; {in single-cell analysis, greater biological insight can be gained by taking into account different views on the same population of cells, such as RNA expression or chromatin conformation \cite{yang2021}.}

Leveraging  information from multiple views can boost  performance. In multi-view learning,  the information can be fused in different ways,  {e.g.}, early fusion and late fusion \cite{zhao2017multi}. In typical early fusion, the data from different views are fused before training,  e.g.,  concatenation of features  \cite{lin2014feature}. In late fusion,  it is commonly the case that multiple sub-models are considered for individual views and  combined to give the final result, e.g., the best single view or the voting committee  \cite{xie2013multi,bekker2015multi}. However, the  mutual influence between different views is not considered sufficiently, and yet is significant to improve the clustering. Various methods that incorporate the couplings of views have been proposed,  {where  a combination of early and late fusions is usually involved} \cite{zhao2017multi}. Co-regularization is a typical technique  {to introduce}  couplings through regularization terms in the loss of the optimization formulation \cite{farquhar2006two,kumar2011,andrew2013deep,Luo2015TensorCC,peng2019}. To  introduce high-order correlations,  tensor-based models can be employed \cite{zhang2015low}, and have been  used mostly in  subspace-based  methods \cite{parsons2004subspace,xie2018unifying} for the consensus low-dimensional subspace. These methods usually reconstruct data from the original view  and deploy different view-specific subspace representations  for the partitioning \cite{zhang2021joint,lv2021multi}, which can risk giving insufficient data descriptions  with linear embeddings  and  view-specific reconstructions. Unlike feature-driven subspace methods,  {graph-based} methods are relation-driven, utilizing point-wise relations between samples \cite{shi2000normalized} and exploring the representation of the data embedded in the given graphs by eigenvalue decomposition of the Laplacian matrix \cite{alzate2008multiway,xia2014robust}. With multiple views, view-specific graphs are employed and integrated to access the  clustering underlying these graphs\cite{zhou2007spectral,nie2017auto,2017cybernetics,zhan2018multiview,li2022multiview}. 
The aforementioned methods have shown satisfactory clustering performance. However, they are optimized with specific alternating iterative procedures that require substantial memory and computations and whose convergence is not always easy to show. By far, much attention has been focused on improving the  accuracy, and yet little effort has been spared on the interpretation and analysis of the shared clustering representation  underlying different views.
Moreover, these methods lack flexibility to infer unseen data, i.e.,  out-of-sample extensions, which in contrast can be implemented in our method. Although the problem of out-of-sample extension was discussed in \cite{li2015large}, it requires external techniques, such as the $k$-nearest  {neighbors}, to process the unseen data 
to locate salient points and then   {propagate} their
labels. 



 The task of spectral clustering can be formulated as a weighted Kernel Principal Component Analysis (KPCA) problem under the Least Squares Support Vector Machine (LSSVM)  framework \cite{suykens2002least}, i.e., Kernel Spectral Clustering (KSC) \cite{alzate2008multiway}, which provides primal-dual insights into the learning
scheme. With KSC, nonlinear feature mappings can be applied to capture complex intrinsic relations of data and meanwhile the point-wise  relations of graph-based information is also utilized to promote the clustering.  Correspondingly, Multi-view KSC (MvKSC) has been proposed in \cite{houthuys2018multi}, which reformulates the weighted KPCA  in  primal with LSSVM and only considers pair-wise correlations in the objective.  MvKSC has shown  favorable accuracy  and less computational cost compared to many state-of-the-art methods, but the computational complexity increases along with the number of views. 
The Restricted Kernel Machine (RKM) \cite{suykens2017deep} is a novel framework that aims to find synergies between kernel methods and neural networks, as, starting from  LSSVM, it gives a  representation consisting of visible and hidden units similar to the Restricted Boltzmann Machine (RBM) \cite{hinton2006}. While RKM has been successfully applied in disentangled representations  \cite{tonin2021unsupervised2,tonin2021unsupervised},  {generative models \cite{pandey2021generative,pandey2022disentangled},} and classification  \cite{houthuys2021tensor}, no previous study has investigated the use of  RKM  or utilized the latent space for spectral clustering  and its   {interpretation} analysis.

\begin{figure}[t]
\begin{center}
\includegraphics[width=0.8\columnwidth]{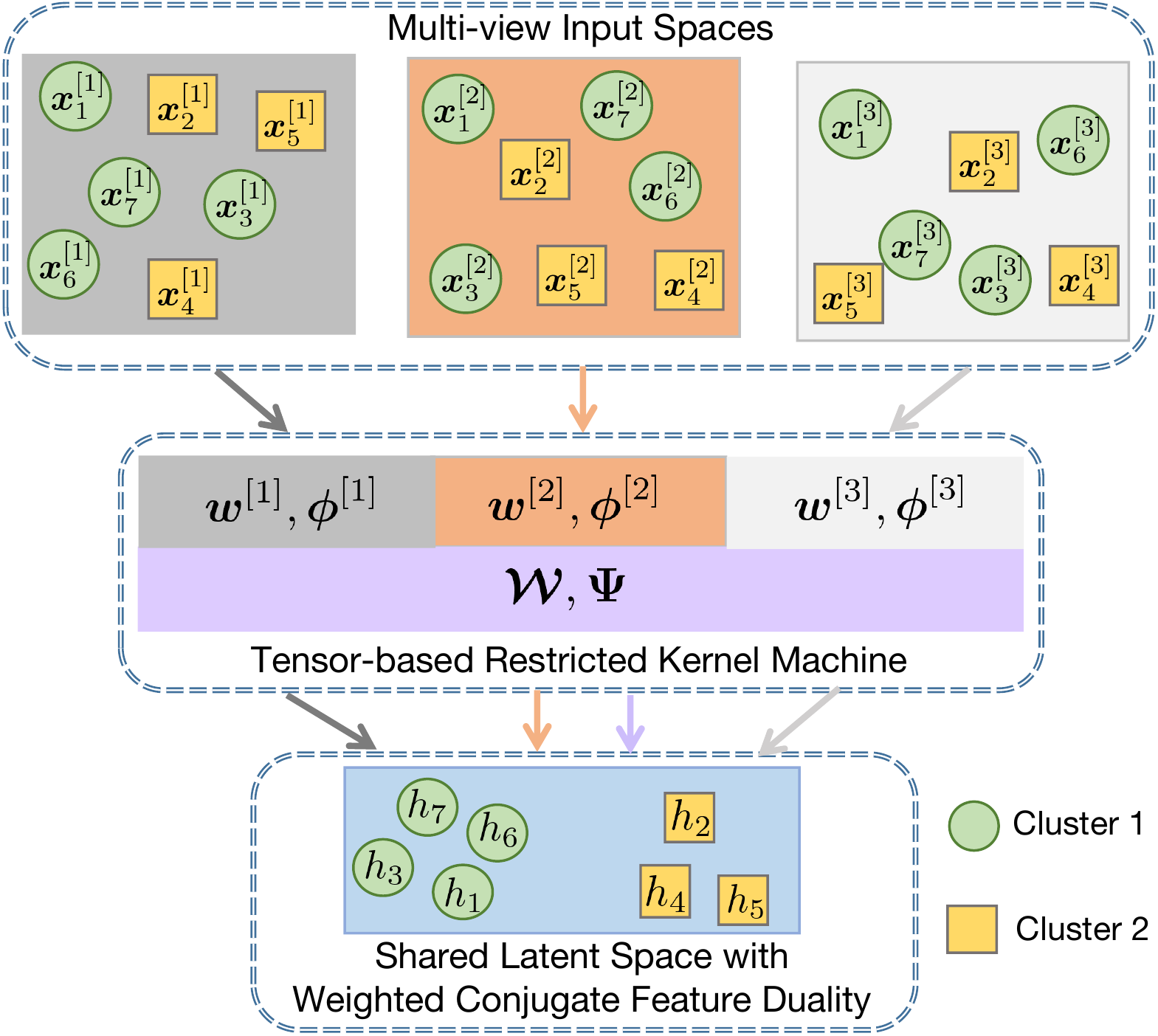}
\caption{Basic diagram of the proposed method. Data $\bm x^{[v]}_i$ are mapped into different feature spaces  using  $\bm w^{[v]}$ and $\bm \phi^{[v]}$ for each view $v=1,2,3$, and meanwhile tensors $\mathcal W$ and $\bm \Psi$ integrating all views are modelled, where view-specific information (in grey, orange, and light grey) is retained and  high-order correlations between different views are incorporated (in purple). Through the proposed weighted conjugate feature duality with RKM, a shared latent space  is attained for spectral clustering, where clusters are well separated and  data interpretation analyses are facilitated via the hidden features common to all views. 
} \label{fig:our:method:intro}
\vspace{-0.5cm}
\end{center}
\end{figure}

In this paper, a novel method is established for  MvSC by developing a modified weighted conjugate feature duality, {bringing}  effective and more interpretable MvSC together with efficient and  simple optimization procedures. In the proposed method,  we derive the objective as an upper bound to the KSC problem in terms of the dual variables employing our constructed conjugate feature duality in  the  RKM framework.  The couplings are achieved  {via} learning projections from the spaces of different views onto a common latent space by sharing the dual variables among all views. 
To  capture higher-order correlation, a Tensor-based RKM model (TMvKSCR) is  constructed by simultaneously integrating  all views into tensor representations, as exemplified in  Figure \ref{fig:our:method:intro}. In the algorithmic aspect, the solution is obtained by a single eigenvalue decomposition,  {whose dimension is  independent} of  the number of views. This  computational efficiency is due  to the  shared latent space in RKM, which eliminates the need to  expand  {the}   dual variables  for multiple views  and meanwhile does not require iteratively alternating parameter optimization,  reducing memory and time consumption.  Besides, our method can be flexibly implemented with   out-of-sample extensions for unseen data, which is of particular interest for large-scale cases:  {during} training with the fixed-size kernel scheme, only $m$ ($m\ll N$ and possibly even $m < \overline{d}$) samples are needed and the training computational complexity reduces from $\mathcal{O}(VN^2\overline{d}+N^2(k-1))$ to $\mathcal{O}(Vm^2\overline{d}+m^2(k-1)+VNm\overline{d})$, where $N$ denotes the number of samples in each view, $\overline{d}$ is the average dimensionality of the views, $k$ is the number of clusters, and $V$ is the number of views. Numerical experiments verify the effectiveness of the proposed method in both accuracy and efficiency. Varied  discussions {and interpretation analysis} are also elaborated on {the common latent space in} our method, e.g., a  sharp eigenvalue decay  in capturing more informative   components and   distinct  latent variable  distributions in well separating the clusters. 
The  contributions of this work mainly include: 
\begin{itemize}
    \item We propose a novel  method for MvSC {based on} the Fenchel-Young inequality, {giving}  an upper bound to the weighted KPCA problem for spectral clustering and {introducing} dual variables  in a different way {that leads} to an  efficient representation and simple optimization.
    \item As the dual variables play the role of conjugated hidden features in the latent space, we impose shared dual variables on all views to construct a common latent space, realizing the view couplings and  facilitating  interpretation analysis   with  dual representations.
    \item The  {resulting} optimization problem only needs  {the solution of} a single eigenvalue decomposition, whose dimension is  independent of the number of views. A tensor-based model  is  established to  capture high-order correlations of the views without increasing  computational complexity.
    \item We discuss the applicability of our method to predict unseen data based on the eigenspace found in training, which also enables to efficiently tackle large-scale data. Numerical experiments verify the superiority of our method in  accuracy, efficiency, and interpretation. 
\end{itemize}

The remainder of this paper is organized as follows: Section \ref{sec:background} summarizes the background of MvSC and RKM. In Section \ref{sec:mv:ksc:rkm}, the proposed method is presented in detail, including representation models together with derivations, the optimization problem, and  discussions. Section \ref{sec:experiments} presents the experimental results, verifying the  advantages of the proposed method  from different aspects. In Section \ref{sec:conclusion},  brief concluding remarks are given.

\section{Background}\label{sec:background}
\subsection{Related Work for MvSC}\label{sec:ksc}
In multi-view learning for spectral clustering (MvSC), data of different views provide complementary information about a common data partitioning underlying all the views. Simply concatenating  the attributes of all views with early fusion  or combining the single-view learning results with late fusion can give reasonable results, but the performance  {is} far from satisfactory due to the lack of utilizing view couplings \cite{parsons2004subspace,zhao2017multi,ma2019}. To this end, great efforts have been made on exploring the common intrinsic clustering structure underlying multiple views with varied methodologies.

The co-regularization  technique is one of the pioneering  methods for MvSC \cite{cai2011,kumar2011,houthuys2018multi}, where extra regularization items are introduced to the optimization objective in training. In this way, the final model is trained by simultaneously considering the view-specific modelling parts and also their mutual influence. For example, \cite{kumar2011} (Co-reg) deploys a co-regularized loss with either a  pairwise or a centroid coupling scheme, and it is commonly considered as one of the  baselines;  \cite{houthuys2018multi} (MvKSC) incorporates pair-wise coupling items with pre-defined weights  to the primal objective with LSSVM.
Graph-based methods have attracted great popularity for MvSC. They aim to learn a common graph containing clustering information via multiple graphs across views \cite{zhou2007spectral,nie2017auto,2017cybernetics,zhan2018multiview}.
As a representative method, \cite{peng2019} (COMIC) jointly considers  the geometric consistency  and the cluster assignment consistency, where the former learns a connection graph for data from the same cluster while the later minimizes the discrepancy of pairwise connection graphs from different views. More recently, \cite{li2022multiview} (SFMC) efficiently coalesces multiple view-wise graphs, learns the weights, and manipulates the joint graph by a connectivity constraint indicating clusters and the selected anchors. Even though one of the main issues tackled in  both COMIC and SFMC is the  parameter-free tuning, there still exist settings that should be tuned in order to well cluster varied datasets, e.g., the threshold of estimating clusters in COMIC.
Despite the diverse fusion techniques in the aforementioned works, higher-order correlations across  views can  be further enhanced by using tensor learning. The representative method based on tensor singular value decomposition is \cite{xie2018unifying} (tSVD-MSC), which adopts self-representation coefficients of views to construct the tensor and uses the  tensor nuclear norm to learn the optimal subspace by rotating  the constructed tensor. It  improved clustering performance over many matrix-based methods, but the computational cost increases  distinctively.
 
Recently, {several  late fusion} MvSC methods have been proposed with diversified fusion techniques as well as  {reduced} computational complexity for large-scale data\cite{wang2019multi}. These methods seek an optimal
partition by combing linearly-transformed partitions
in individual views and using $k$-means  for the learned
consensus partition matrix to
get cluster labels. In 
\cite{liu2021one} (OP-LFMVC),  learning of the consensus partition matrix  and   generation of cluster labels  {are unified} by developing a four-step alternate algorithm. Similarly, in \cite{liu2021one-iccv} (OPMC) a one-pass fusion method  {is proposed}, which unifies and jointly optimizes the steps of  matrix factorization and partition generation, where  an alternate strategy  is constructed to obtain the parameters.  {Additionally,
t}owards large-scale data and computational efficiency, bipartite  graphs are effective strategies, as only part of the samples need to be selected to build the graphs in relation to the data structure for clustering. In  \cite{li2015large} ({MVSCBP}), local manifold fusion is used to integrate heterogeneous features and approximate the similarity measures using bipartite graphs with selected salient points, and  out-of-sample inference is  also considered by label propagations, further improving the efficiency. In  SFMC, bipartite  graphs are also used to build sample connectivity by selecting anchors \cite{li2022multiview}.

KSC \cite{alzate2008multiway} formulates  spectral clustering  as a weighted KPCA problem  with LSSVM\cite{suykens2002least}, where the clustering information is included based on the eigendecomposition of a modified similarity matrix of the data and the random walk model in relation to the graph information is utilized for the weighting \cite{pmlr-vR3-meila01a}. More discussion on the links between spectral clustering and the weighted KPCA can be found in \cite{ham2004kernel,alzate2008multiway}.  {The solution (stationary points) to KSC is given by an eigenvalue decomposition  scaled by $N\times N$. The projections by the weighted KPCA, i.e., clustering scores, are computed {by} the first eigenvectors and then used to encode the clusters.} For multi-view learning, KSC has been   analogously extended to MvKSC  \cite{houthuys2018multi} by imposing additional constraints in the primal model, {which yields a $VN\times VN$ eigenvalue  problem, where $V$ is the number of views.
}

\subsection{Restricted Kernel Machines}\label{sec:rkm}
\begin{figure}[ht!]
\begin{center}
\includegraphics[width=0.3\columnwidth]{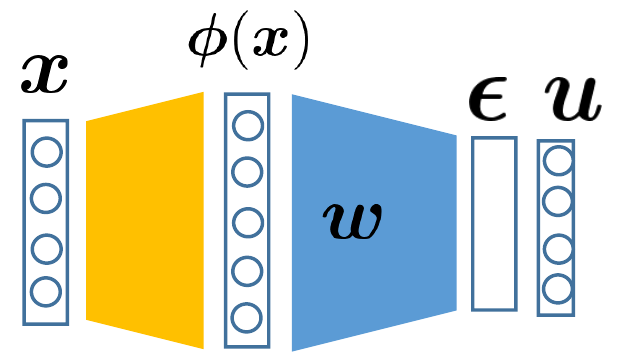}
\caption{Graphical topology of RKM for KPCA.
An input $\bm x$ is mapped to a feature space using a feature map ${\bm  \phi}$ with  projections $\bm \epsilon$  {coupled} with  hidden features $\bm u$ in the latent space.
} \label{fig:rkm}
\end{center}
\end{figure}

The RKM formulation for KPCA, i.e., the unweighted settings for KSC, is given by an upper bound of the objective function in LSSVM   using the Fenchel-Young inequality  \cite{rockafellar1974conjugate}:
$
\frac{1}{2\lambda}\bm \epsilon^T\bm \epsilon+ \frac{\lambda}{2}\bm u^T\bm u \geq \bm \epsilon^T\bm u, \forall \bm \epsilon, \bm u \in \mathbb R^s,
$ which introduces the hidden features $\bm u$ and leads to the conjugate feature duality. The resulting KPCA objective  is given by:
\begin{equation} \label{eq:obj:ksc:rkm}
    \mathcal J
       = - \sum\limits_{i=1}^N \bm \epsilon_i^T \bm u_i + \dfrac{\lambda}{2} \sum\limits_{i=1}^N {\bm u_i}^T{\bm u_i} + \dfrac{\eta}{2} \mathrm{Tr}\left(\bm W^T\bm W\right),
\end{equation}
where   the hidden features $\bm u_i$ for each sample are conjugated to the projections $\bm \epsilon_i =\bm W ^T\bm \phi(\bm x_i)$ along all $s$ directions with the interconnection weight  $\bm W \in \mathbb R^{d_h \times s}$ and  feature map $\bm \phi: \mathbb R^d \rightarrow \mathbb R^{d_h}$, and $\eta, \lambda \in \mathbb R_{+}$ are the regularization constants. 
Here, the first summation term resembles the energy function of  RBM  \cite{hinton2006}, involving connections between visible units in the input space and hidden units in the latent space, as  demonstrated in Figure \ref{fig:rkm}. 



%
%

By characterizing the stationary points of $\mathcal{J}$ in $\eqref{eq:obj:ksc:rkm}$,  the following  eigenvalue  problem is obtained:
\begin{equation}\label{eq:dual:kpca:rmk}
    \frac{1}{\eta}{\bm \Omega_{\rm c}}\bm U^T = \bm U^T \bm \Lambda,
\end{equation}
where  $\bm U = [\bm u_1, \ldots, \bm u_N]\in \mathbb R^{s\times N}$ and $\bm \Lambda = {\rm diag} \{\lambda_1, \allowbreak \ldots, \allowbreak \lambda_N\}$ with $s\leq N$ selected principal components, and  ${\bm \Omega_{\rm c}}$ is the centered Mercer kernel matrix \cite{vapnik1999overview} induced by ${\bm \phi}(\cdot)$. One can verify that the solutions corresponding to different eigenvectors and  eigenvalues in \eqref{eq:dual:kpca:rmk} all lead to  $\mathcal{J}=0$  \cite{suykens2017deep}.






 \section{Multi-view Spectral Clustering via Shared Latent Space with RKM}
\label{sec:mv:ksc:rkm}
In this section, the model formulation and the optimization of the proposed method are derived. Then, the decision rule for encoding the clusters and the out-of-sample extension are explained.

\subsection{Model Formulation}\label{sec:model}


\subsubsection{Modified Weighted Conjugate Feature Duality for Spectral Clustering}\label{sec:appendix:weighted:duality}
Under weighted least squares settings, the conjugate feature duality for KPCA with $s$ components in \eqref{eq:obj:ksc:rkm} can be extended based on the following inequality: 
\begin{equation}\label{eq:f-y:appendix}
\frac{1}{2\lambda}\bm \epsilon^T\bm P\bm \epsilon+ \frac{\lambda}{2}\bm u^T\bm P^{-1}\bm u \geq \bm \epsilon^T\bm u, \quad \forall \bm \epsilon, \bm u \in \mathbb R^s,
\end{equation}
with  $\bm P \succ 0$ and $\lambda >0$ \cite{pandey2020robust},
where the hidden features $\bm u$ {regarding each sample} are conjugated to the projections $\bm \epsilon$. It can be seen that  {the use of} matrix $\bm P \in \mathbb R^{s\times s}$  realizes the weighting over the selected $s$  components.

{In KSC-based methods, spectral clustering can be formulated as a weighted KPCA problem with LSSVM. The corresponding primal model for the single-view KSC with $k$ clusters is expressed as\cite{alzate2008multiway}:}
\begin{equation}\label{eq:kpca:lssvm:appendix}
    \begin{array}{lll}
    \min\limits_{\bm w^{{(l)}}, \bm e^{{(l)}} } & \mathcal{J}_{\rm KSC} =  \dfrac{\eta}{2} \sum\limits_{l=1}^{k-1}{\bm w^{{(l)}}}^T\bm w^{{(l)}} -
  \sum\limits_{l=1}^{k-1}\dfrac{1}{2\lambda^{(l)}}{\bm e^{{(l)}}}^T\bm D^{-1}\bm e^{(l)} \\
   {\hspace{6mm}\rm{s.t.}} & \bm e^{{(l)}} = \left(\bm \Phi - \mathbf{1}_{N}\hat{\bm \mu}^{T}\right)\bm w^{(l)},
\end{array}
\end{equation}
where $\bm w^{{(l)}}, \bm e^{{(l)}} \in \mathbb R^N$, $l = 1, \dots, k-1, \lambda^{(l)} >0$, $\bm 1_{N}$ denotes a vector with $N$ ones, and $\hat{\bm \mu}$ is used to center the data. The matrix $\bm D\in \mathbb R^{N\times N}$ is the so-called degree matrix constructed in relation to the random walks model \cite{pmlr-vR3-meila01a} and is formulated as a positive definite diagonal matrix with  $D_{ii} = \sum\nolimits_{j}K(\bm x_j, \bm x_i)$ and $K: \mathbb R^d \times \mathbb R^d \mapsto \mathbb R$ being the kernel function. {For simplicity, we use ${\bm \Phi}_{\rm c}=[{{\bm \phi}_{\rm c}(\bm x_1)}, \ldots, {{\bm \phi}_{\rm c}(\bm x_N)}]^T$ to denote  the centered feature map $\bm \Phi - \mathbf{1}_{N}\hat{\bm \mu}^{{[v]}^T}$ in the paper.}
The score variables  $\bm e^{{(l)}}$  {encode the  data clusters}. Note that a set of $k-1$ binary clustering indicators are sufficient to {differentiate} $k$ clusters, so that $l = 1, \dots, k-1$.


{
In order to tackle spectral clustering, we firstly develop a modified version of weighted conjugate feature duality, which is essential in constructing the objective function of KSC with RKM. Differently, we introduce the hidden features  through  the Fenchel-Young inequality, but apply it to the $N$-dimensional vector of score variables $\bm e=[e_1, \dots, e_N]^T$ with the  weighting matrix $\bm D$, where the conjugation is conducted along each projection direction of all samples}, such that 
\begin{equation}
\frac{1}{2\lambda}\bm e^T\bm D^{-1}\bm e+ \frac{\lambda}{2}\bm h^T\bm D\bm h \geq \bm e^T\bm h, \quad \forall \bm e, \bm h \in \mathbb R^N,
\end{equation}
{where the matrix $\bm D \in \mathbb R^{N\times N}$ achieves the weighting over all $N$ samples to build point-wise connectivity between samples.}
Then, the upper bound objective  $\overline{\mathcal{J}}_{\rm KSC}$ is obtained as:
\begin{equation}\label{eq:wkpca:ksc}
\begin{split}
 \mathcal{J}_{\rm KSC} =&  \dfrac{\eta}{2} \sum\limits_{l=1}^{k-1}{\bm w^{{(l)}}}^T\bm w^{{(l)}} -
\sum\limits_{l=1}^{k-1}  \dfrac{1}{2\lambda^{(l)}}{\bm e^{{(l)}}}^T\bm D^{-1}\bm e^{(l)}  \\
  \leq &
  -\sum\limits_{l=1}^{k-1}{\bm e^{{(l)}}}^T\bm h^{(l)} +  \sum\limits_{l=1}^{k-1}  \dfrac{\lambda^{(l)}}{2}{\bm h^{(l)}}^T\bm D\bm h^{(l)}
  \\
   &+\dfrac{\eta}{2} \sum\limits_{l=1}^{k-1}{\bm w^{{(l)}}}^T\bm w^{{(l)}}\triangleq \overline{ \mathcal{J}}_{\rm KSC},
  \end{split}
\end{equation}
{with conjugated hidden features  $\bm h^{(l)} = [h_1^{(l)}, \ldots, h_N^{(l)}]^T$, {$l=1, \ldots, k-1$}. Thus, the derived $\overline{\mathcal{J}}_{\rm KSC}$ in \eqref{eq:wkpca:ksc}} formulates the corresponding RKM objective $\overline{ \mathcal{J}}_{\rm KSC}$ for the  weighted KPCA settings in tackling spectral clustering. By substituting the constraints $\bm e^{{(l)}}=\bm \Phi_{\rm c}\bm w^{(l)}$ into the RKM objective $\overline{ \mathcal{J}}_{\rm KSC}$, we have
\begin{equation}
\begin{split}
    \overline{ \mathcal{J}}_{\rm KSC} = & -\sum\limits_{l=1}^{k-1}{\left(\bm \Phi_{\rm c}\bm w^{(l)} \right)}^T\bm h^{(l)}
    +  \sum\limits_{l=1}^{k-1}\dfrac{\lambda^{(l)}}{2}{\bm h^{(l)}}^T\bm D\bm h^{(l)}
    \\
    &  +\dfrac{\eta}{2} \sum\limits_{l=1}^{k-1}{\bm w^{{(l)}}}^T\bm w^{{(l)}},
 \end{split}
\end{equation}
where the first summation term in this RKM establishes the interconnections between visible units from the input space and the hidden units from the latent space.

{For multi-view learning with data $\mathcal{D}^{[v]}=\{\bm x_i^{[v]} \}_{i=1}^N, v=1, \ldots, V$, the view-specific RKM objective $\mathcal{J}^{[v]}_{\rm RKM}$ of our proposed model in primal is formulated as:}
\begin{equation}
\begin{split}  
 \overline{ \mathcal{J}}^{[v]}_{\rm KSC}  =& -\sum\limits_{l=1}^{k-1}{\left(\bm \Phi^{[v]}_{\rm c}\bm w^{[v]^{(l)}}\right)}^T\bm h^{{[v]}^{(l)}} +
    \sum\limits_{l=1}^{k-1}\dfrac{\lambda^{{(l)}}}{2}{\bm h^{{[v]}^{(l)}}}^T\bm D^{[v]} \bm h^{{[v]}^{(l)}}\\
     & + \dfrac{\eta}{2}\sum\limits_{l=1}^{k-1}{\bm w^{[v]^{(l)}}}^T\bm w^{[v]^{(l)}} \triangleq \mathcal{J}^{[v]}_{\rm RKM},
\end{split}
\end{equation}
{with  score variables (projections) in \eqref{eq:kpca:lssvm:appendix} expanded as}
\begin{equation}\label{eq:e:decision}
    \bm e^{[v]^{(l)}} = (\bm \Phi^{[v]} - \mathbf{1}_{N}\hat{\bm \mu}^{{[v]}^T})\bm w^{[v]^{(l)}} \triangleq \bm \Phi^{[v]}_{\rm c}\bm w^{[v]^{(l)}},
\end{equation}
where $\bm h^{{[v]}^{(l)}}\in \mathbb R^{N}$ are the conjugated hidden features {in the induced latent space of the $v$-th view}, $\bm \Phi^{[v]} = [\bm \phi^{[v]}(\bm x^{[v]}_1), \ldots, \bm \phi^{[v]}(\bm x^{[v]}_N)]^T \in  \mathbb R^{N\times d_h}$ contains the view-specific feature maps $\bm \phi^{[v]}: \mathbb R^{d^{[v]}} \rightarrow \mathbb R^{d_h^{[v]}}$ with interconnection weight $\bm w^{[v]^{(l)}} \in \mathbb R^{d_h^{[v]}}
$, {and $ \bm \Phi^{[v]}_{\rm c}$ denotes the corresponding centered feature map}. Analogously, $\bm D^{[v]} \in \mathbb R^{N\times N}$  is the degree matrix for the $v$-th view. 

In the constructed RKM objective  $\mathcal{J}^{[v]}_{\rm RKM}$ of KSC, the first two sets of summation terms give the view-specific connections between the inputs and the hidden units of the latent space.


 


\subsubsection{Shared Latent Space}\label{sec:mv:ksc:rkm:matrix}
Under the RKM learning framework, we start from  weighted KPCA and introduce a modified  conjugate feature duality to formulate the  RKM objective  $\mathcal{J}^{[v]}_{\rm RKM}$ for spectral clustering in each view, as shown in Section \ref{sec:appendix:weighted:duality}.  For  multiple views, we  simultaneously incorporate the view-specific RKM objectives of KSC, leading to the  objective:
\begin{equation}\label{eq:obj:mv:ksc:rkm}
    \begin{split}
     \mathcal{J}_{\rm Mv} =  &  \sum\limits_{v=1}^V \mathcal{J}^{[v]}_{\rm RKM}. 
    \end{split}
\end{equation}

\begin{figure}[t]
\begin{center}
\includegraphics[width=0.5\columnwidth]{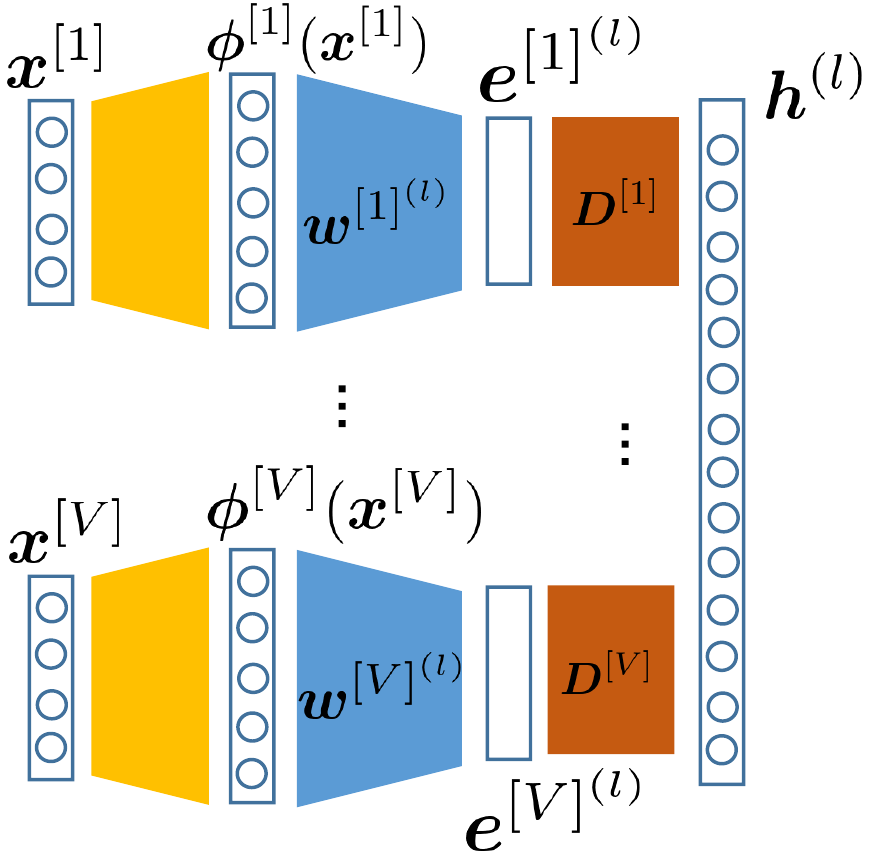}
\caption{Graphical topology of the shared latent space with RKM. Each view $v$ is mapped to a feature space using a feature map ${\bm \phi}^{[v]}$ with view-specific projections ${{\bm e}^{[v]}}^{(l)}$, {where the degree matrix $\bm D^{[v]}$ is in relation to the random walks model promoting the clustering}. The couplings between views are obtained by deploying the  conjugated hidden features ${\bm h}^{(l)}$ in the common latent space, shared by all views.} \label{fig:mv:ksc:rkm}
\end{center}
\end{figure}

Importantly, we impose the couplings of different views by sharing  hidden features in the 
latent space, due to the conjugate dual variables being hidden features in RKM. Thus, in the context of this paper, the sharing of hidden features in the induced common latent space is achieved  by
\begin{equation}
\bm h^{{[1]}^{(l)}}   = \cdots = \bm h^{{[V]}^{(l)}}  \triangleq \bm h^{{(l)}}.
\end{equation}
 Figure \ref{fig:mv:ksc:rkm} gives the graphical topology and shows that  the learned features from the spaces of different views are projected into the common latent space, which well retains the view-specific information and meanwhile jointly considers the mutually shared information of all views.  {The constructed common latent space perfectly fits and  facilitates the goal of our task,
i.e., finding the common data partitions for all views, as we can learn a shared representation of the data from all views and promote the clustering in the latent space. Analysis can also be simplified for all views,  providing a more straightforward way for interpretation and data
discovery in such common latent space  with dual representations.}

\subsubsection{Tensor-based modelling}\label{sec:mv:ksc:tensor}
Though the shared hidden features in the latent space can lead to the couplings of all views,  the feature maps $\bm \phi^{[v]}$ and their interconnection weights $\bm w^{{[v]}^{(l)}}$ are still calculated separately for  individual views, as shown in Figure \ref{fig:mv:ksc:rkm}. To boost high-order correlations, we incorporate tensor learning to our model, which extends the fusion strategy and further explores the complementary information  across  views.

In the objective, we introduce  multi-order weight tensors and  rank-1 feature mapping tensors  to build the model, instead of using view-specific  matrices and vectors.
Thus,  the full objective  using  tensor representations is attained as:
\begin{equation}\label{eq:tmv:ksc:rkm:objective}
\begin{split}
    \mathcal{J}_{\rm TMv} = &
    - \sum\limits_{l=1}^{k-1}\sum\limits_{i=1}^N \left<{{\bm \Psi}}_{{\rm c}, i},  {\mathcal{W}}^{(l)} \right >h^{{(l)}}_i + \\
  &   \sum\limits_{v=1}^V\sum\limits_{l=1}^{k-1}\dfrac{\lambda^{(l)}}{2}{\bm h^{{(l)}}}^T\bm D^{[v]} \bm h^{{(l)}}
      +\dfrac{\eta}{2}  \sum\limits_{l=1}^{k-1} \left<\mathcal{W}^{(l)}, \mathcal{W}^{(l)}\right >, 
\end{split}
\end{equation}
where $\mathcal{W}^{(l)}\in \mathbb R^{d_h^{{[1]}}\times\cdots \times d_h^{{[V]}}}$ are $V$-th order tensors of the interconnection weights {for the $l$-th projection}, ${\bm \Psi}_{{\rm c}, i} \in \mathbb R^{d_h^{{[1]}}\times\cdots \times d_h^{{[V]}}}$ are rank-1 tensors composed by the outer products of all view-specific feature maps, such that ${\bm \phi}_{\rm c}^{[1]}(\bm x_i^{[1]}) \otimes \cdots \otimes {\bm \phi}_{\rm c}^{[V]}(\bm x_i^{[V]})$. The  outer product  and the  tensor inner product are calculated by $
     \mathcal{A}_{{m_1}\ldots {m_P}} \coloneqq \bm a_{1_{m_1}} \otimes \cdots \otimes \bm a_{P_{m_P}}$ and $\left<\mathcal{A}, \mathcal{B}\right> \coloneqq \sum\nolimits_{m_1 = 1}^{M_1} \cdots \sum\nolimits_{m_p = 1}^{M_P} \mathcal{B}_{{m_1}\ldots {m_P}}\mathcal{A}_{{m_1}\ldots {m_P}}$, respectively,
with $P$ vectors $\bm a_1\in \mathbb R^{M_1}, \ldots, \bm a_P\in \mathbb R^{M_P}$ for the outer product and 
two $P$-th order tensors  $\mathcal A, \mathcal B \in \mathbb R^{M_1\times \cdots \times M_P}$ for the inner product. Note that, as $\bm h^{(l)}$ is shared by all views, the calculations between $\bm h^{(l)}$ and degree matrix $\bm D^{[v]}$ in the second summation term remain the same.

\begin{figure}[t]
\begin{center}
\includegraphics[width=0.55\columnwidth]{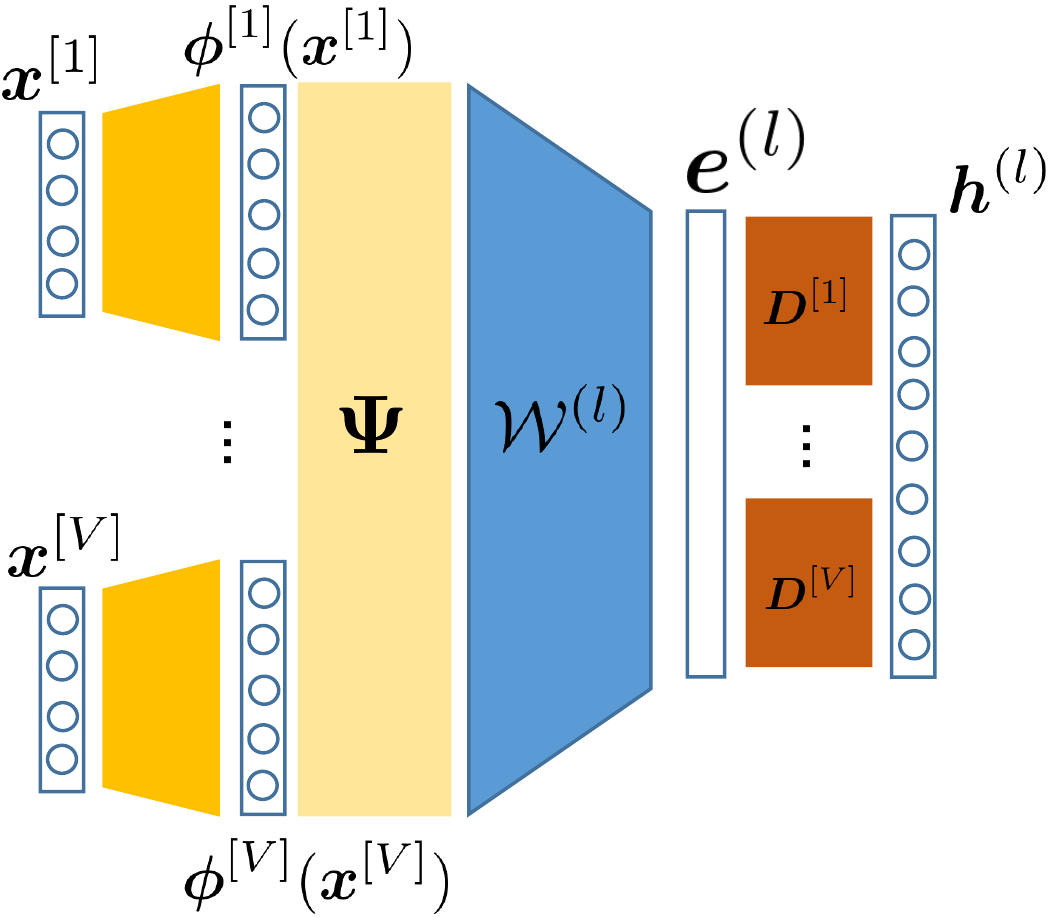}
\caption{Graphical topology of the tensor-based model. All views are mapped to a  feature tensor  ${\bm \Psi}$ constructed by the outer product of the  feature maps ${\bm \phi}^{[v]}$  of all views. Then, the fused information is  connected by the weight tensors $\mathcal{\bm W}^{(l)}$ with the conjugated hidden features ${\bm h}^{(l)}$ in the common latent space.
}\label{fig:mv:ksc:rkm:tensor}
\end{center}
\end{figure}

 {The constructed  tensors ${\bm \Psi}_{{\rm c}, i}$ and  $\mathcal{W}^{(l)}$ join the feature maps and interconnection weights, respectively, of all views together.}
Figure \ref{fig:mv:ksc:rkm:tensor} shows the graphical topology of the proposed tensor learning, which incorporates the view-specific feature maps within the tensors and performs the calculations of all views simultaneously, leading to the tensor-based high-order couplings and information fusion in an earlier stage compared to Figure \ref{fig:mv:ksc:rkm}, which, by contrast,  fuses the views  in a relatively later stage. 


With the shared latent space and  tensor learning, we then generalize  the proposed model, namely {TMvKSCR}, extending to  weighted views and integrated fusions. Firstly, we integrate the matrix (later fusion in Figure \ref{fig:mv:ksc:rkm})  and the tensor (earlier fusion in Figure \ref{fig:mv:ksc:rkm:tensor}) representations  by combining $\mathcal{J}_{\rm Mv}$ and $\mathcal{J}_{\rm TMv}$,  for diversified couplings and fusions.
Secondly, we consider a more general case where the relative impacts of different views can be varied, for example the reliability (noise) varies in different views. Thus,  the  clustering scores from 
different views are weighted in $ \mathcal{J}^{[v]}_{\rm RKM}$  by replacing $\bm e^{{[v]}^{(l)}}$  with $\sqrt{\kappa^{[v]}}\bm e^{{[v]}^{(l)}}$, where the yielded  objective is denoted $\mathcal{J}_{\rm RKM}^{{\kappa}^{[v]}}$. Note that the weighted outer product involving the tensors in  $\mathcal{J}_{\rm TMv}$  is equivalent to an amplification {magnitude} of  $\sqrt{\kappa^{[1]}\cdots\kappa^{[V]}}$ without changing the solution space.

Accordingly,  our model generalized in this paper, i.e., {TMvKSCR}, is given by the following objective:
\begin{equation}\label{eq:obj:general}
    \mathcal{J}_{\rm TMvR} = \varrho \sum\limits_{v=1}^V \mathcal{J}_{\rm RKM}^{{\kappa}^{[v]}} + (1-\varrho)\mathcal{J}_{\rm TMv}.
\end{equation}
In  our model,   {we learn projections of multi-view data} into the common latent space by sharing the  {conjugated} dual variables {(hidden features)} in RKM, and the tensor learning further  boosts high-order correlations across views, incorporating the global inter-view information. At the same time,  the view-specific information is well retained by the individual feature maps of weighted views, incorporating the  local intra-view information. Therefore, the intra-view relations and the inter-view correlations are both explored. 

\subsection{Problem Optimization} \label{sec:appendix:general}
In the optimization, instead of tackling the explicit feature maps $\bm \phi^{[v]}$ in  {in primal representations}, we consider characterizing the stationary points of $\mathcal{J}_{\rm TMvR}$ with conjugate feature duality \cite{suykens2017deep} and the kernel trick \cite{mercer1909}.

By taking the partial derivatives to the weights and the hidden features,  the  conditions of  the stationary points  for  $\mathcal{J}_{\rm TMvR}$ are  characterized by:
\begin{equation}
    \left\{
    \begin{split}
    \dfrac{\partial \mathcal{J}_{\rm TMvR} }{\partial \bm  w^{[v]^{(l)}}}=&  \eta \bm w^{[v]^{(l)}} - \sqrt{\kappa^{[v]}}\sum\limits_{i=1}^N{\bm  \phi}_{\rm c}^{[v]}\left(\bm x_i^{[v]}\right)h_i^{(l)} = 0,    \\
     \dfrac{\partial \mathcal{J}_{\rm TMvR}}{\partial \mathcal{W}^{(l)}_{m_1 \ldots m_V}}=&  \eta \mathcal{W}^{(l)}_{m_1 \ldots m_V} - \sum\limits_{i=1}^N \prod\limits_{v=1}^V{\bm  \phi}_{\rm c}^{[v]}\left(\bm x_i^{[v]}\right)_{m_v}h_i^{(l)} = 0,  \\
     & \text{with} \  m_v = 1, \ldots, d^{[v]}_h,\\
    \dfrac{\partial\mathcal{J}_{\rm TMvR}}{\partial  h_i^{(l)}} =& \varrho\sum\limits_{v=1}^V\sqrt{\kappa^{[v]}}{\bm w^{[v]^{(l)}}}^T{\bm  \phi}_{\rm c}^{[v]}\left(\bm x_i^{[v]}\right)  \\
    &+ (1-\varrho)\left<{{\bm \Psi}}_{{\rm c}, i},  \mathcal{\bm W}^{(l)} \right >  - \sum\limits_{v=1}^V\lambda^{(l)}D_{ii}^{[v]}h_i^{(l)} =  0, \\
    & \text{with} \ i=1, \ldots, N,
    \end{split}\label{eq:kkt:general}
    \right.
\end{equation}
 {for $l=1, \ldots, k-1$.} From the first two equations in \eqref{eq:kkt:general}, it follows:
\begin{equation}
    \left\{
    \begin{array}{ll} 
     \bm w^{[v]^{(l)}}=\dfrac{1}{\eta} \sqrt{\kappa^{[v]}}\sum\limits_{i=1}^N{\bm  \phi}_{\rm c}^{[v]}\left(\bm x_i^{[v]}\right)h_i^{(l)},    \\
      \mathcal{W}^{(l)}_{m_1 \ldots m_V}  =\dfrac{1}{\eta} \sum\limits_{i=1}^N \prod\limits_{v=1}^V{\bm  \phi}_{\rm c}^{[v]}\left(\bm x_i^{[v]}\right)_{m_v}h_i^{(l)}.
    \end{array}
    \right.
\end{equation}
Eliminating $\bm w^{[v]^{(l)}}$  and $\mathcal{W}^{(l)}_{m_1 \ldots m_V}$  gives the following  {expression} in the conjugated features $h^{(l)}_i$,
\begin{equation} 
\begin{split}
   &   \dfrac{1-\varrho}{\eta} \prod\limits_{v=1}^V\sum_{m_v=1}^{d_h^{[v]}}{\bm  \phi}_{\rm c}^{[v]}\left(\bm x_i^{[v]}\right)_{m_v} \left(\sum\limits_{j=1}^N \prod\limits_{v=1}^V{\bm  \phi}_{\rm c}^{[v]}\left(\bm x_j^{[v]}\right)_{m_v}h_j^{(l)}\right)\\
&+  \dfrac{\varrho}{\eta} \sum\limits_{v=1}^V{\kappa^{[v]}}\sum\limits_{j=1}^{N}{\bm \phi}_{\rm c}^{[v]}\left(\bm x_j^{[v]}\right)^T{\bm \phi}^{[v]}_{\rm c}\left(\bm x_i^{[v]}\right)h_j^{(l)} = \lambda^{(l)}\sum\limits_{v=1}^VD_{ii}^{[v]}h_i^{(l)}, \\
& {\text{for} \  l= 1, \ldots, k-1, \ \text{and} \  i = 1, \ldots, N,}
\end{split}
\end{equation}
yielding  the following eigenvalue problem \eqref{eq:dual:second:extension:general}  {for} the optimization of our method
\begin{equation}\label{eq:dual:second:extension:general}
\begin{split}
  & \dfrac{1}{\eta}\left(\sum\limits_{v=1}^V\bm D^{[v]}\right)^{-1}\left(\varrho \sum\limits_{v=1}^V  \kappa^{[v]} {\bm \Omega}_{\rm c}^{[v]} + \left(1-\varrho\right)\bigodot_{v=1}^V {\bm \Omega}_{\rm c}^{[v]}\right)\bm H\\
  & \hspace*{3.5cm} = \bm H \bm \Lambda,
\end{split}
\end{equation}
where 
${\bm \Omega}_{\rm c}^{[v]} =  {\bm \Phi}_{\rm c}^{{[v]}}{{\bm \Phi}_{\rm c}^{{[v]}}}^T$ denotes the centered kernel matrix induced by the kernel function $K^{[v]}: \mathbb{R}^{d^{[v]}} \times \mathbb{R}^{d^{[v]}} \rightarrow \mathbb{R}$, and  $\bm H\coloneqq [\bm h^{(1)}, \ldots,\bm h^{(k-1)}]\in \mathbb R^{N\times(k-1)}$ contains the selected $k-1$  components (eigenvectors) corresponding to the largest $k-1$ eigenvalues $\bm \Lambda\coloneqq {\rm diag} \{\lambda^{(1)}, \allowbreak \ldots, \allowbreak \lambda^{(k-1)} \}$.  {By solving \eqref{eq:dual:second:extension:general}, the first $k-1$ eigenvectors are  used to calculate the score variables $\bm e^{{[v]}^{(l)}}$ for clustering. Note that 
$\eta$ is a scaling factor  without affecting the solution space of the optimization problem and thereby set  as 1.}


It can be seen from \eqref{eq:dual:second:extension:general} that,  {although the proposed model integrates both view-wise information and high-order correlations across all views into a common latent space with dual representations, the resulting} optimization  is simple in {formulation}, i.e., solving an $N\times N$  eigenvalue decomposition  {problem for} the first $k-1$ eigenvectors,  which gives the stationary points of the objective  {function $\mathcal{J}_{\rm TMvR}$ as derived in \eqref{eq:kkt:general}. Thus, the optimization procedure of our method does not need to iteratively alternate updating model parameters, which instead is commonly required} in many related works including subspace and graph-based methods, e.g., \cite{li2015large,peng2019,liu2021one,zhang2021joint,lv2021multi,liu2021one-iccv,li2022multiview}. Moreover, the scale of \eqref{eq:dual:second:extension:general}  is independent of the number of views $V$, and this efficiency owes to the shared hidden features of all views with the conjugate feature duality in RKM, so that there is no  need to expand the size of dual variables along with the increasing  number of views. 

{The} clustering score variables $\bm e^{[v]^{(l)}}$ in \eqref{eq:e:decision} can  {then} be computed using the kernel trick in dual  {form}. Accordingly,  the dual model representation for the score variables corresponding to each data point is calculated by
\begin{equation}\label{eq:calculate:score}
    \hat e^{{[v]}^{(l)}}_i(\bm x^{[v]}_i)   = \sum\limits_{j=1}^N h^{(l)}_jK^{[v]}(\bm x_i^{[v]}, \bm x_j^{[v]}).
\end{equation}
With the $k-1$ eigenvectors obtained from \eqref{eq:dual:second:extension:general}, i.e., $\bm h^{(l)}=[h_1^{(l)}, \ldots, h_N^{(l)}]^T$ for $l=1, \ldots, k-1$, the score variables of all data points are obtained and their cluster labels can be encoded accordingly, as explained in  Section \ref{sec:decision:testing}.



\subsection{Decision Rule and Out-of-Sample Extension}\label{sec:decision:testing}
The encoding vector of a certain sample  from multiple views   $\{\bm x_i^{[v]}\}_{v=1}^V$ consists of the  indicators:
\begin{equation}\label{eq:train:encoding}
     {\rm sign}\left(e_i^{{[v]}^{(1)}}\right), \dots, {\rm sign}\left(e_i^{{[v]}^{(k-1)}}\right), 
 \end{equation}
 where 
 the score variables are calculated in the dual form by 
    $\bm e^{{[v]}^{(l)}} = {\bm \Omega}_{\rm c}^{[v]}\bm h^{(l)}$ in \eqref{eq:calculate:score}. The $k$ most occurring encoding vectors  form the codebook. In multi-view cases, the clustering assignment can be done in two ways: individual assignment and ensemble assignment.
 For the former, the cluster assignment can be done separately for each view, and $V$ codebooks are created, i.e., $\mathcal C^{[v]}=\{c_p^{[v]}\}_{p=1}^k$. These clustering results can vary  across views. For the latter, which we use in this paper, a single cluster assignment is conducted for all views by
$
        \bm e^{(l)}_{\rm mean} = \sum\nolimits_{v=1}^V\beta^{[v]}\bm e^{{[v]}^{(l)}}$, where only one codebook $\mathcal C=\{c_p\}_{p=1}^k$ is created and  $\beta^{[v]}$  can be simply taken as $1/V$  or  calculated
 separately. Each data point  is  {then} assigned to the cluster of the closest codeword based on the ECOC decoding \cite{dietterich1994solving}.  

The proposed method can be flexibly applied to cluster unseen data, i.e., the out-of-sample extension,  a  merit from KSC-related techniques \cite{alzate2011out}.
Similar to \eqref{eq:calculate:score}, the  scores for  unseen data $\mathcal{D}^{[v]}_{\rm test}=\{{\bm x^{[v]}_{\rm test}}_j\}_{j=1}^{N_{\rm te}}$ can be calculated to correspond to the projections onto the eigenvectors found in the training, i.e.,
\begin{equation}\label{eq:out:of:sample}
    \bm e^{{[v]}^{(l)}}_{\rm test} = {\bm \Omega}_{\rm c, test}^{[v]}{\bm h^{(l)}},
\end{equation}
where ${\bm \Omega}_{\rm c, test}^{[v]}\in \mathbb R^{N_{\rm te}\times N}$ is the centered testing kernel matrix  computed from
$\bm \Omega_{{\rm test}_{ij}}^{[v]} = K^{[v]}(\bm x_{{\rm test}_i}^{[v]}, \bm x_j^{[v]})$.
Hence, we do not need to re-conduct the training procedure   when new data are included, which is particularly useful in dealing with large-scale data. For instance, the fixed-size kernel scheme \cite{mall2013} can be applied in our method using  dual representations, as a small subset of data is selected for the training, and then the clusters  of the complete dataset can be inferred by using out-of-sample extensions.
Algorithm 1 details the training and the out-of-sample
extension of the proposed method.

\begin{algorithm}[t]
   \caption{ {Training and  Out-of-sample Inference Procedures of the Proposed TMvKSCR}}
   \label{alg:1}
\begin{algorithmic}[1]
   \STATE {\bfseries Input:} $\mathcal{D}^{[v]}=\{\bm x_i^{[v]} \}_{i=1}^N$ and $\mathcal{D}^{[v]}_{\rm test}=\{{\bm x^{[v]}_{\rm test}}_j\}_{j=1}^{N_{\rm te}}$ (if present), $K^{[v]}(\cdot, \cdot)$, $k$, $\kappa^{[v]}$ and $\varrho$;
    \STATE {\bfseries Output:} Clusters of data. 
   \STATE Compute the centered kernel matrices ${\bm  \Omega}_{\rm c}^{[v]}$ and their degree matrices  $\bm D^{[v]}$; 
   \STATE 
   Solve the eigenvalue problem in \eqref{eq:dual:second:extension:general} and obtain $\bm h^{(l)}$; 
   \STATE 
  Compute the score variables $\bm e^{{[v]}^{(l)}}$ by \eqref{eq:calculate:score} and build the  {codebook $\mathcal C =\{c_p\}_{p=1}^k$}, with $c_p\in \{-1, 1\}^{k-1}$;
   \STATE Assign each  $\bm x_i^{[v]}$ to its cluster by applying the codebook $\mathcal C$ on   {${\rm sign}(\bm e^{{[v]}^{(1:k-1)}}_i)$ of \eqref{eq:train:encoding}};
   \  $\triangleright$ Clusters of $\mathcal{D}^{[v]}$
   \IF{$\mathcal{D}^{[v]}_{\rm test} \neq \emptyset$}
    \STATE  Compute the centered kernel matrices ${\bm  \Omega}_{\rm c, test}^{[v]}$; 
\STATE Compute the score variables $\bm e_{\rm test}^{{[v]}^{(l)}}$   {by \eqref{eq:out:of:sample}};
\STATE Assign each ${\bm x^{[v]}_{\rm test}}_j$ to its cluster by applying codebook $\mathcal C$ on   {${\rm sign}(\bm e^{{[v]}^{(1:k-1)}}_i)$ of  \eqref{eq:train:encoding}};  \ $\triangleright$ Clusters of $\mathcal{D}_{\rm test}^{[v]}$
   \ENDIF
   
\end{algorithmic}
\end{algorithm}

\subsection{Computational Complexity Analysis} 


 {
The computation involved in the proposed TMvKSCR mainly covers two aspects, i.e., obtaining the kernels  and calculating the {first} $k-1$ eigenvectors in \eqref{eq:dual:second:extension:general}. Denoting the average input dimensions of all views as $\overline{d}=(1/V)\sum_{v=1}^V d^{[v]}$, the first aspect of  kernel computation has a time complexity of $\mathcal{O}(VN^2\overline{d})$\cite{houthuys2018multiclassification}, which is commonly involved in spectral clustering methods  {when calculating the similarity matrices consisting of the point-wise relations  between  samples}. The second aspect of solving the first $k-1$ eigenvectors in \eqref{eq:dual:second:extension:general} gives a time complexity of $\mathcal{O}((k-1)N^2)$, which is independent of the number of views $V$ thanks to the shared latent space and leads to the storage complexity  as $\mathcal{O}(N^2)$.
}

 {Employing} the out-of-sample extension, unseen data can be predicted and analogously the fixed-size kernel scheme can be applied for large-scale data. We assume that $m$ samples are used in the training. The time complexity of the aforementioned two aspects  is reduced to $\mathcal{O}(Vm^2\overline{d})$ and $\mathcal{O}((k-1)m^2)$, respectively, with $m\ll N$ and possibly even $m< \overline{d}$.   Then, the out-of-sample extension to predict the whole dataset needs the complexity of $\mathcal{O}(VmN\overline{d})$.  In this case, the maximal storage complexity is only $\mathcal{O}(m^2)$.  Thus, the main computation is approximately quadratic in the number of samples used for training. Using fixed-size kernels schemes with $m\ll N$ ({possibly with $m < \overline{d}$}), the proposed method can efficiently deal with large-scale data. Note that  our method does not alternate to update parameters with multiple iterations, so it has a fixed complexity, which further helps efficiency.



\section{Numerical Experiments}\label{sec:experiments}
In this section, numerical experiments are conducted to evaluate the proposed method,  with comparisons to other related methods on various datasets. Experiments are implemented on MATLAB
2020a/Python 3.7  with 64 GB RAM and a 3.7 GHz Intel i7 processor.  {More experimental details  are supplemented in  Appendix \ref{appendix:experiment:setup}, and the source code can be downloaded from  \url{https://github.com/taralloc/mvksc-rkm}}. 

\subsection{Experimental Setups}\label{sec:test:setup}

\textbf{Datasets} To evaluate the proposed method, two synthetic datasets (Synth 1/2) are presented, where samples are generated by  two-dimensional Gaussian mixture models. 
For high-dimensional real-world data, we consider the multi-view benchmarks of 3Sources \cite{greene2009matrix}, Reuters \cite{liu2013multi},  Ads  \cite{kushmerick1999,Luo2015TensorCC}, Image Caption (ImgC) \cite{kolenda2002}, YouTube Video Games (YT-VG) \cite{madani2012}, and  NUS-WIDE (NUS) \cite{Chua2009NUSWIDEAR}. The large-scale Reuters (L-Reuters)  \cite{Dua:2019}  is then used to evaluate  the out-of-sample extension and the large-scale case, as introduced in Table \ref{tab:real:data:info}, where  more details of the datasets are given in Appendix \ref{appendix:experiment:setup}.


\begin{table}[ht!]
\caption{Description of datasets  {used for the experiments.}}
\begin{tabular}{lcccl}
\toprule
  Dataset    & $N$ & $V$ & $k$ & $d^{[1:V]}$ \\
  \hline
  Synth 1 & 1000 & 3 & 2 & 2,2,2 \\
 Synth 2 & 1000 & 2 & 2 & 2,2 \\
  3Sources & 169  & 3 & 6 &  3560, 3631, 3068 \\
  Reuters & 600  & 3 & 6 &  9749, 9109, 7774 \\
  Ads       & 3279 & 3 & 2 &  587, 495, 472 \\
  ImgC & 1200  & 3 & 3 &  768, 192, 3522 \\
  YT-VG & 2100  & 3 & 7 &  1000, 512, 2000 \\
  NUS & 5970 & 5 & 10 &  64, 225, 144, 73, 128 \\
  L-Reuters & 18758 & 5 & 6 &  21531, 24892, 34251, 15506, 11547\\
\bottomrule
\end{tabular}
\label{tab:real:data:info}
\end{table}

\noindent \textbf{Evaluation metrics and compared methods}  {To assess the clustering performances of different methods,  Adjusted Rand index (ARI) \cite{hubert1985comparing} and  Normalized Mutual Information (NMI) \cite{strehl2002cluster} are widely used to measure the clustering consistency between the true labels and the predicted ones. The ARI metric takes values in $[-1,1]$, with 0 indicating a random partitioning and 1 indicating the perfect clustering  identical to the ground truth. The NMI metric normalizes the cluster entropy to give the mutual information between the obtained clustering and the ground-truth clustering. It takes values in $[0,1]$, where 1 indicates the perfect clustering.}

 {We evaluate our proposed TMvKSCR with comparisons to different related methods for MvSC. Firstly, the late fusion based on the best view of view-specific KSC (best) and the early fusion based on  attribute concatenation in KSC (concat) are tested. Then, several representative and more recent MvSC methods are compared, including {Co-reg} \cite{kumar2011}, {MVSCBP}\cite{li2015large}, {tSVD-MSC} \cite{xie2018unifying}, {MvKSC}\cite{houthuys2018multi}, {COMIC}\cite{peng2019}, {OP-LFMVC}\cite{liu2021one}, {OPMC}\cite{liu2021one}, and {SFMC}\cite{li2022multiview}, details of which are in Section \ref{sec:ksc}.}
Regarding hyperparameter selection, we tune the hyperparameters of each tested method by grid search in the ranges suggested by the authors in their papers. Though some methods provide a parameter-free setting by using heuristically estimated values, some hyperparameters still need to be tuned to achieve desirable performance on the datasets used in the experiments, such as the clustering threshold  in  COMIC,  the kernels in OP-LFMVC, and the number of salient points and nearest  {neighbors} in MVSCBP. Thus, we also tune such hyperparameters or use the suggested values in the papers of these compared methods and report the obtained optimal results for each dataset.
 The shared hyperparameters among all methods are tuned under the same settings, e.g.,  kernel parameters of the RBF bandwidth, and the method-specific hyperparameters, e.g.,  $\kappa^{[v]}$ and $\varrho$ in our method, are tuned in separate ranges. More details of the setups can be found in Appendix \ref{appendix:experiment:setup}.


\subsection{Performance Comparisons}\label{sec:overall:results}

%
\subsubsection{Clustering Performance}
\begin{figure}[ht!]
	\centering
		\includegraphics[width=0.115\textwidth]{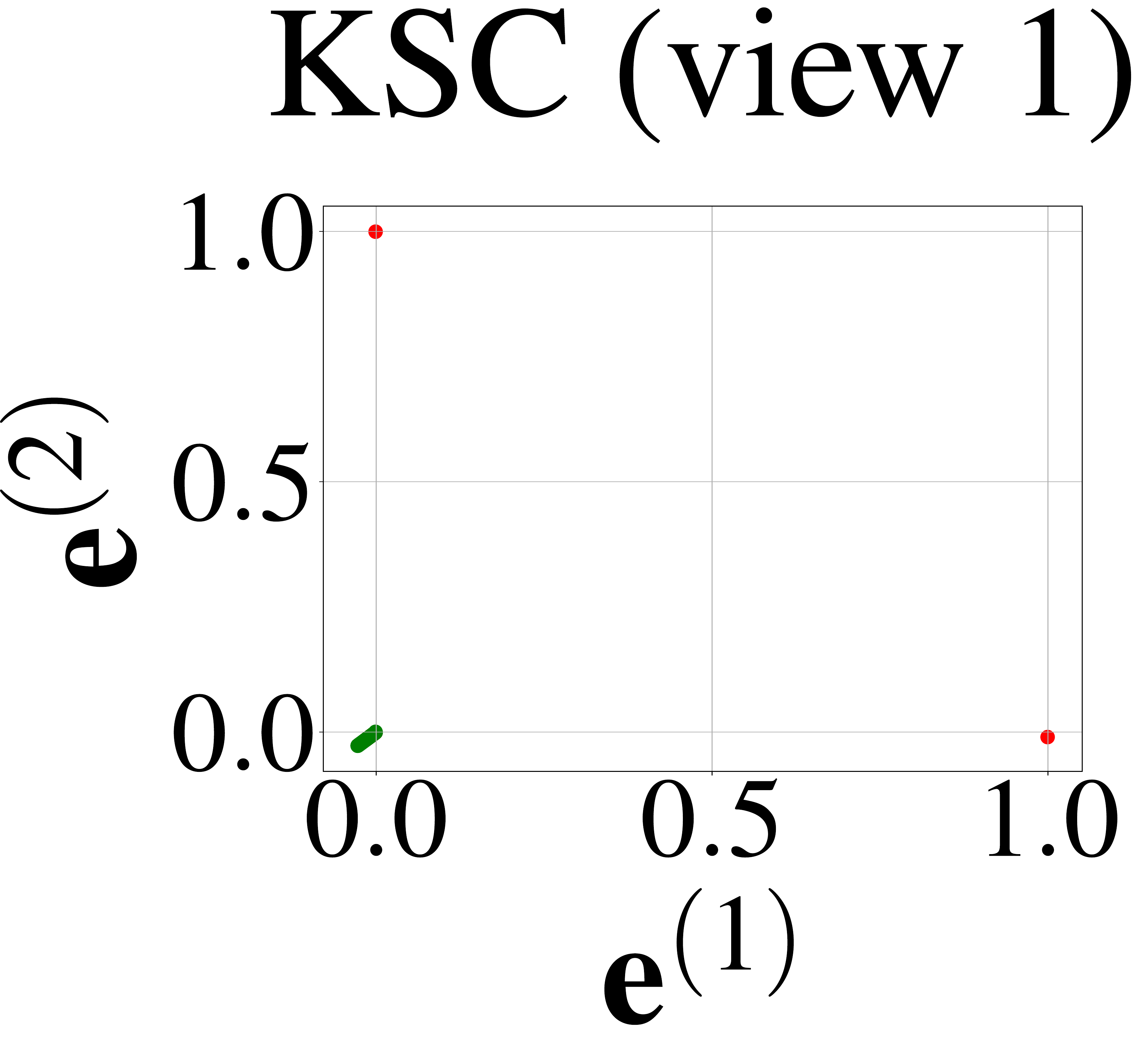}
		\includegraphics[width=0.115\textwidth]{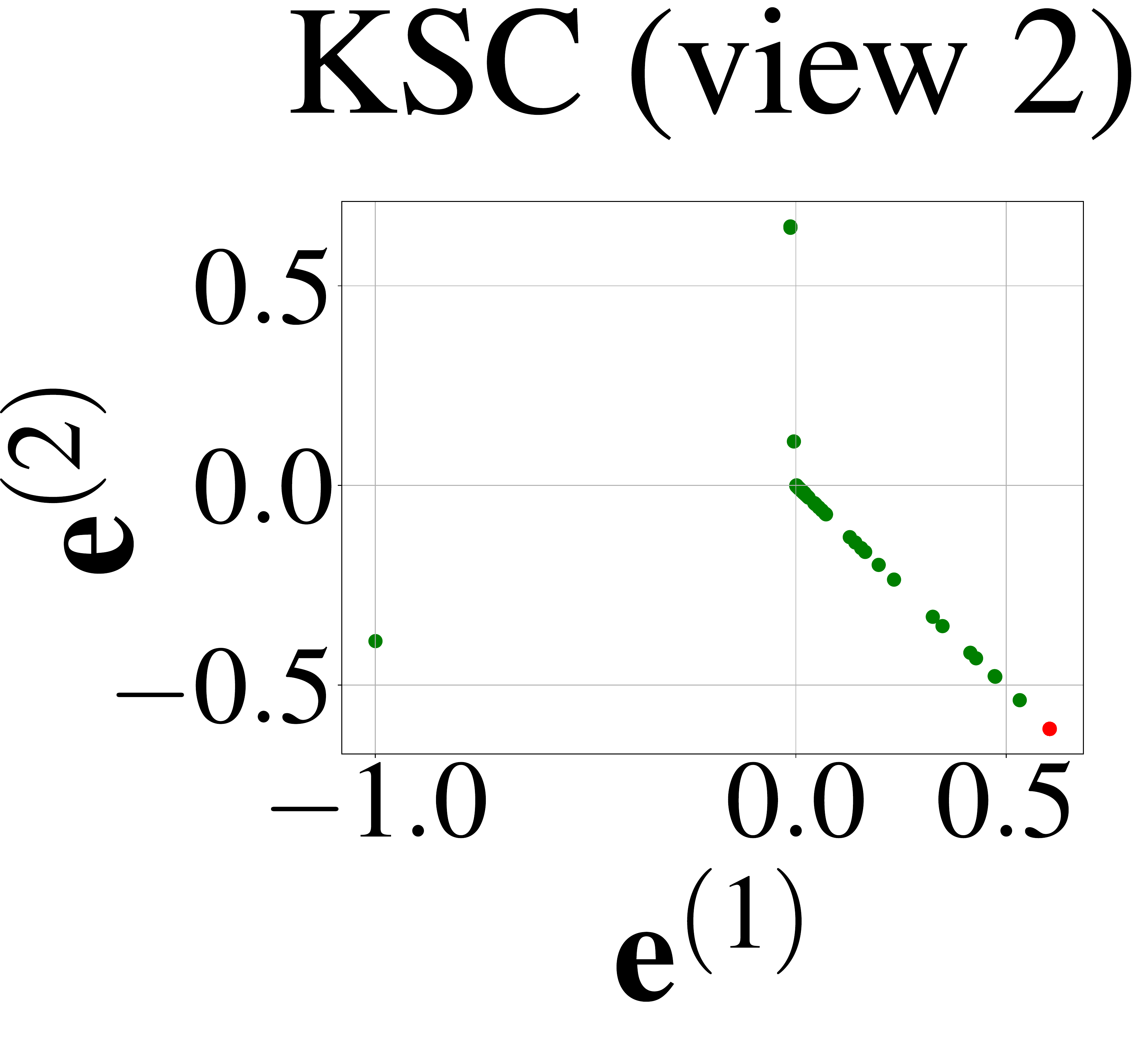}
		\includegraphics[width=0.115\textwidth]{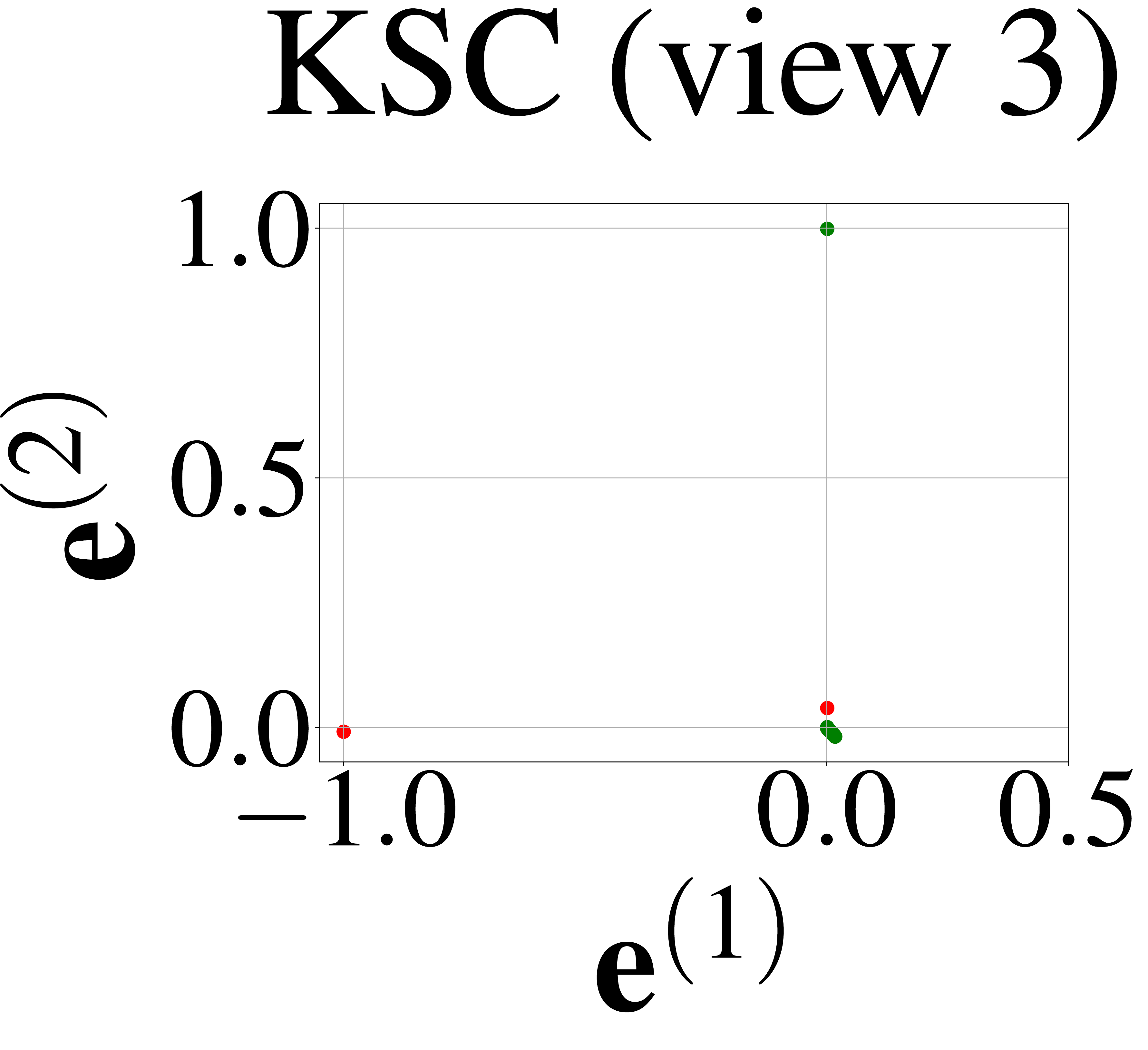}
		\includegraphics[width=0.115\textwidth]{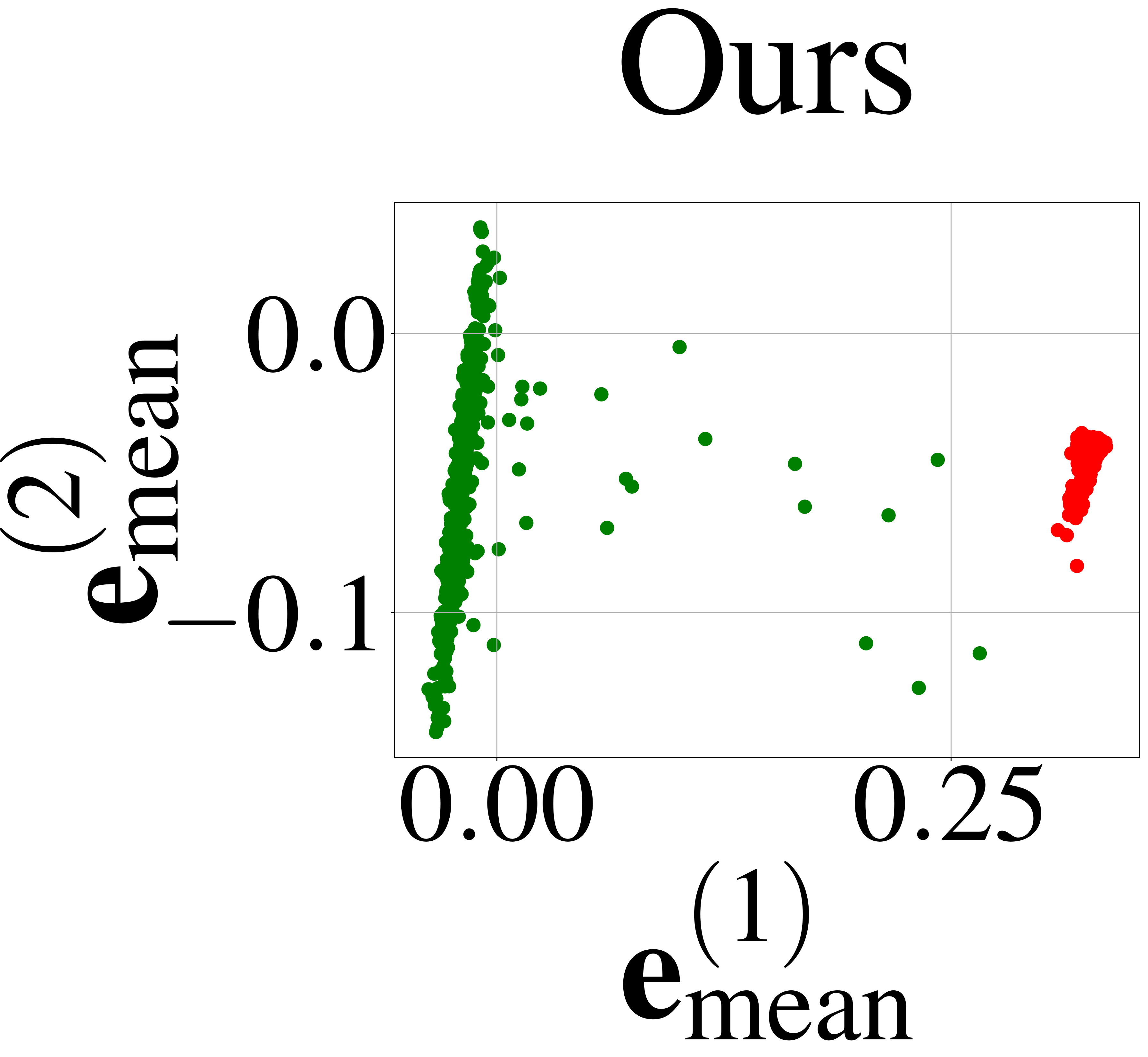}
	\caption{Score variables for Synth 1 with RBF  ($\sigma^2=0.05$).}
	\label{fig:synth2}
\end{figure}

 {As a simple illustration,} Figure  \ref{fig:synth2} plots the score variables of our method and of  KSC on the synthetic dataset Synth 1, where the colors represent the two clusters. It shows that the projections of KSC
applied  to each view cannot find a good separation, while  the two groups found by our method are well separated.  {The improved separation between the clusters leads to better quantitative} clustering performance, as shown in Tables \ref{tab:best:ari} and \ref{tab:best:nmi}.

\begin{table*}[ht!]
    \caption{Clustering performance comparisons in terms of the best ARI. } 
    \centering
\begin{tabular}{lllllllll}
\toprule
Method & Synth 1 & Synth 2 & 3Sources & Reuters & Ads & ImgC & YT-VG &  NUS \\
\midrule
KSC (best)                                      & 0.678  & 0.336  & 0.647 & 0.194 & 0.433 & 0.434 & 0.030 & 0.026 \\
KSC (concat)                                     & 0.996 & 0.316 & 0.688 & 0.217    & 0.178 & 0.351 & 0.259 & 0.046 \\
Co-Reg & 0.996& 0.324&0.655&0.219&0.272& 0.471& 0.024& 0.056\\
%
tSVD-MSC                                        & 0.817 & 0.002 & 0.690& 0.248    & -0.001 & 0.637 & 0.396 & 0.050 \\
$\text{MvKSC}$ & 0.996 & 0.507 & 0.642 & 0.233& 0.433 & 0.794& 0.205 & 0.050 \\
%
COMIC                                              & 0.168 & 0.072 & 0.489 & 0.073   & 0.037 & 0.511 & 0.055 & 0.037 \\
OP-LFMVC & 0.988 & 0.064 & 0.649 & 0.231 & 0.241 & 0.772 & 0.388 & 0.056 \\
OPMC & 0.972  & 0.315  & 0.321 & 0.197 & 0.326 & 0.430 & 0.239 & 0.040 \\
SFMC & 0.968  & 0.236  & 0.377 & 0.105 & 0.223 & 0.578 & 0.406 & 0.021 \\
MVSCBP & 0.996 & 0.322  & 0.698 & 0.244 & 0.366 & 0.609 & 0.336 & 0.050 \\
$\text{TMvKSCR}$ & \textbf{1.000}&   \textbf{0.568} & \textbf{0.717}& \textbf{0.273} & \textbf{0.463} & \textbf{0.884}& \textbf{0.526} & \textbf{0.070} \\
\bottomrule
\end{tabular}
    \label{tab:best:ari}
\end{table*}

\begin{table*}[ht!]
    \caption{Clustering performance comparisons in terms of the best NMI. } 
    \centering
\begin{tabular}{lllllllll}
\toprule
Method & Synth 1 & Synth 2 & 3Sources & Reuters & Ads & ImgC & YT-VG &  NUS \\
\midrule
KSC (best)                                      & 0.639  & 0.289  & 0.614 & 0.295 & 0.344 & 0.478 & 0.051 & 0.033 \\
KSC (concat)                                     & 0.989 & 0.161 & 0.681 & 0.311    & 0.143 & 0.385 & 0.324 & 0.072 \\
Co-Reg   & 0.989& 0.273& 0.688& 0.347& 0.215& 0.541& 0.050& 0.095 \\
%
tSVD-MSC                                        & 0.737 & 0.001 & 0.750& 0.353    & 0.039 & 0.674 & 0.467 & 0.080 \\
$\text{MvKSC}$ & 0.989& 0.373&0.669&0.318& 0.333& 0.739& 0.285& 0.071 \\
%
COMIC                                              & 0.116 & 0.166 & 0.600 & 0.272   & 0.127 & 0.548 & 0.402 & 0.095 \\
OP-LFMVC & 0.970 & 0.071 & 0.645 & 0.350 & 0.097 & 0.748 & 0.431 & 0.087 \\
OPMC & 0.939  & 0.259  & 0.477 & 0.320 & 0.154 & 0.525 & 0.329 & 0.083  \\
SFMC & 0.936  & 0.206  & 0.469 & 0.388 & 0.223 & 0.627 & 0.654 & \textbf{0.136} \\
MVSCBP & 0.989 & 0.266  & 0.695 & \textbf{0.397} & 0.289 & 0.659 & 0.555 & 0.081 \\
TMvKSCR &\textbf{1.000}& \textbf{0.428} & \textbf{0.756}& 0.350& \textbf{0.350} & \textbf{0.835} & \textbf{0.686}& 0.089  \\
\bottomrule
\end{tabular}
    \label{tab:best:nmi}
\end{table*}

For real-world multi-view  datasets, Table \ref{tab:best:ari}  and Table \ref{tab:best:nmi} give the performance evaluation with comparisons to other baselines and state-of-the-art methods. Our  method distinctively improves over KSC and MvKSC on all considered
datasets regarding both ARI and NMI, showing the effectiveness of our  novel fusion technique  with the RKM framework. 
Compared to other approaches,  {our method achieves the best overall performance on the tested datasets.} More specifically, the ARI attained by our method on ImgC is 0.884, compared to the  {second-best} of 0.794 by MvKSC. On YT-VG, our method with a simple linear kernel gives an approximately 32.8\% improvement in  ARI compared to the second best tSVD-MSC, which also uses tensor learning but is  {considerably} more computationally expensive.  COMIC commonly gives much higher NMI than ARI, and this is likely due to NMI not being adjusted for chance, since
 COMIC automatically determines $k$, which might be overestimated, e.g., it gives 570 clusters for NUS. In such cases, NMI can be very high even with random partitions \cite{romano2016, amelio2017}, but the actual clustering  is  poor.  {Our method also outperforms the recently proposed SFMC, except for slightly lower NMI on Reuters.}  {For NUS, we only use $m=1000$ samples for the training and infer the remaining data using out-of-sample extensions based on the eigenvectors optimized in training, and our method still achieves the highest ARI and the second highest NMI, showing the  effectiveness of the out-of-sample predictions in our method in well balancing accuracy and efficiency}. Overall, our  method gains the highest ARI on all datasets and  maintains the best NMI for most  datasets, verifying the efficacy of the proposed  tensor-based model and  the shared latent space by leveraging  conjugate feature duality.

\subsubsection{Efficiency}
 {Efficiency comparisons are also conducted to evaluate our method. In this experiment, we compare the running time for all tested methods, and list the corresponding results in
 Table \ref{tab:runtime}, where the most efficient multi-view method is underlined.} Thanks to the shared latent space, our method shows similar  running time with single-view methods, i.e., KSC (best/concat), and gives comparable efficiency as OP-LFMVC, which is efficiently designed for large-scale data but shows lower overall accuracy on most datasets in Tables \ref{tab:best:ari} and \ref{tab:best:nmi}. Compared to other  methods, our method is distinctively more efficient. For example, compared to
the tensor-based tSVD-MSC that attains the second best overall performance but is the most computationally expensive,  our method  gives higher  performance in considerably less time.  When
using fixed-size kernel schemes ($m=1000$) with random selection for the larger dataset NUS, our method gains the highest ARI
(0.070) in $\approx$0.5s, which is an order of magnitude
faster than   OP-LFMVC ($\approx$7s).


%
%
%
\begin{table}[ht]
\caption{Average training time of 10 runs (in seconds).}
\centering
\resizebox{\columnwidth}{!}{
\begin{tabular}{llHlllll}
\toprule
Method                                                      & 3Sources & Rtrs1 & Reuters & Ads & ImgC & YT-VG & NUS\\
\midrule
KSC (best)                                      & 0.15 & 1.63  & 0.28  & 2.34  & 1.91  & 1.51  & 9.50  \\
KSC (concat)                                         & 0.17  & 2.25  & 0.28 & 4.28  & 0.52  & 1.46  & 10.52 \\
Co-Reg                                                   & 0.44  & 5.39  & 1.42 & 21.42  & 2.01  & 7.58 & 143.55  \\
%
tSVD-MSC                                                   & 3.61 & 447.87  & 55.17  & 458.48  & 51.53  & 152.35  & 5480.66  \\
$\text{MvKSC}$                                       & 0.23  & 6.50  & 1.68  & 34.08  & 19.26  & 8.15  & 892.76  \\
COMIC                                                       & 1.36  & 182.28  & 19.51 & 17.94  & 8.25  & 18.06  & 100.77 \\
OP-LFMVC & 0.64 & 0.99 & 0.66 & \underline{1.14} & 0.75 & \underline{1.27} & 7.19 \\ 
OPMC & 4.09 & & 12.58 & 7.65 & 4.05 & 15.85 & 30.49 \\
SFMC & 0.556 & & 1.311 & 15.116 & 2.510 & 6.555 & 28.703 \\
MVSCBP & 1.848 & & 2.047 & 9.727 & 3.372 & 4.147 & 9.065 \\
Ours      & \underline{0.14}  & 2.09 & \underline{0.27}  & 4.47  & \underline{0.57}  & 1.58  & \underline{0.50}  \\
\bottomrule
\end{tabular}
}
\label{tab:runtime}
\end{table}

%
%
%
%

\begin{figure}[ht!]
\centering
\
\includegraphics[scale=1]{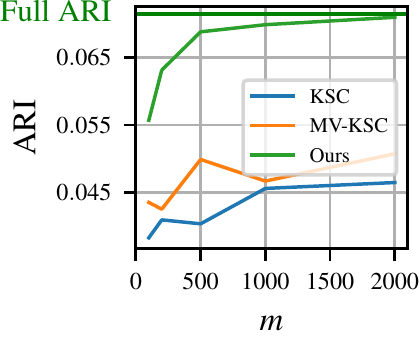}
\includegraphics[scale=1]{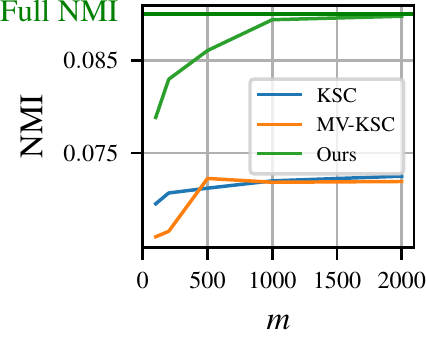}
\caption{ARI and NMI on  {NUS} with varied $m$ averaged over 10 runs. Given the same $m$, MvKSC using Lagrangian duality tackles a $5m\times 5m$  eigenvalue  problem, while our method maintains higher  ARI and NMI by solving an $m\times m$ one instead,  which is of particular benefit  under limited computational resources.}
\label{fig:large-scale}
\end{figure}

As introduced before, our method  can be implemented with  out-of-sample extensions, which greatly benefit  large-scale cases, e.g., selecting a subset $m \ll N$ for the training and then inferring the remaining data. 
  {To further verify these points}, we  compare to the methods, i.e., KSC (concat) \cite{alzate2008multiway}, MvKSC \cite{houthuys2018multi}, {MVSCBP}\cite{li2015large},  that are extendable to unseen data on both   {NUS and} L-Reuters. 
In Figure \ref{fig:large-scale}, we firstly evaluate the out-of-sample extension in the proposed  method on NUS   {with random subset selection, but selection can also be done with other strategies, e.g., using Renyi entropy \cite{girolami2002} or anchor selections \cite{li2015large,li2022multiview}. Our method}  maintains higher  ARI and NMI on varying $m$.  {It can also be observed that our method shows  negligible performance degradation to the full accuracy on NUS},  {when only $1/3$ of the complete data are randomly selected and  used in training, further verifying the informativeness of  hidden features found in our method.}


We further compare with   {the scalable MVSCBP, OP-LFMVC, and SFMC on L-Reuter dataset. Consistent with the settings used in the compared papers \cite{li2015large,li2022multiview}, $m=400$ salient points, or namely the anchors, are used in  building the graphs for training; we use the same $m=400$ salient points in our training phase}. The comparison results on accuracy and efficiency are given in Table \ref{tab:lreuters}.  {It can be seen that our method achieves competitive performance than the compared methods, attaining the highest ARI. It is worth mentioning that our method shows distinctively  less CPU time, where  the running time is reduced by  more than $1\sim 2$ orders of magnitude than the compared scalable methods, showing  the efficiency of the out-of-sample extension in the proposed method and its effectiveness for large-scale cases.}

\begin{table}[ht!]
\centering

\caption{ {Clustering performance and running time (in seconds) on L-Reuters.}}\label{tab:lreuters}
\begin{tabular}{llll}
\toprule
          & ARI & NMI &  Time \\
\midrule
MVSCBP  & 0.270   & 0.339   & 3.88        \\
OP-LFMVC  & 0.257  & 0.317   & 48.54        \\
SFMC & 0.126   & \textbf{0.341}   & 58.15       \\
Ours      & \textbf{0.312}   & 0.312  & \underline{0.29}     \\
\bottomrule
\end{tabular}

\end{table}

\subsection{Further Analysis and Interpretation}\label{sec:test:further}
Leveraging the shared latent space with dual representations in RKM  provides versatile perspectives on clustering interpretation and analysis  {in a more straightforward way.}  {We can  conduct analysis directly on the constructed model itself to reveal the clustering results without external tools}. 

\subsubsection{Eigenvalues}  
\begin{figure}[ht]
\centering
  \includegraphics[width=0.14\textwidth]{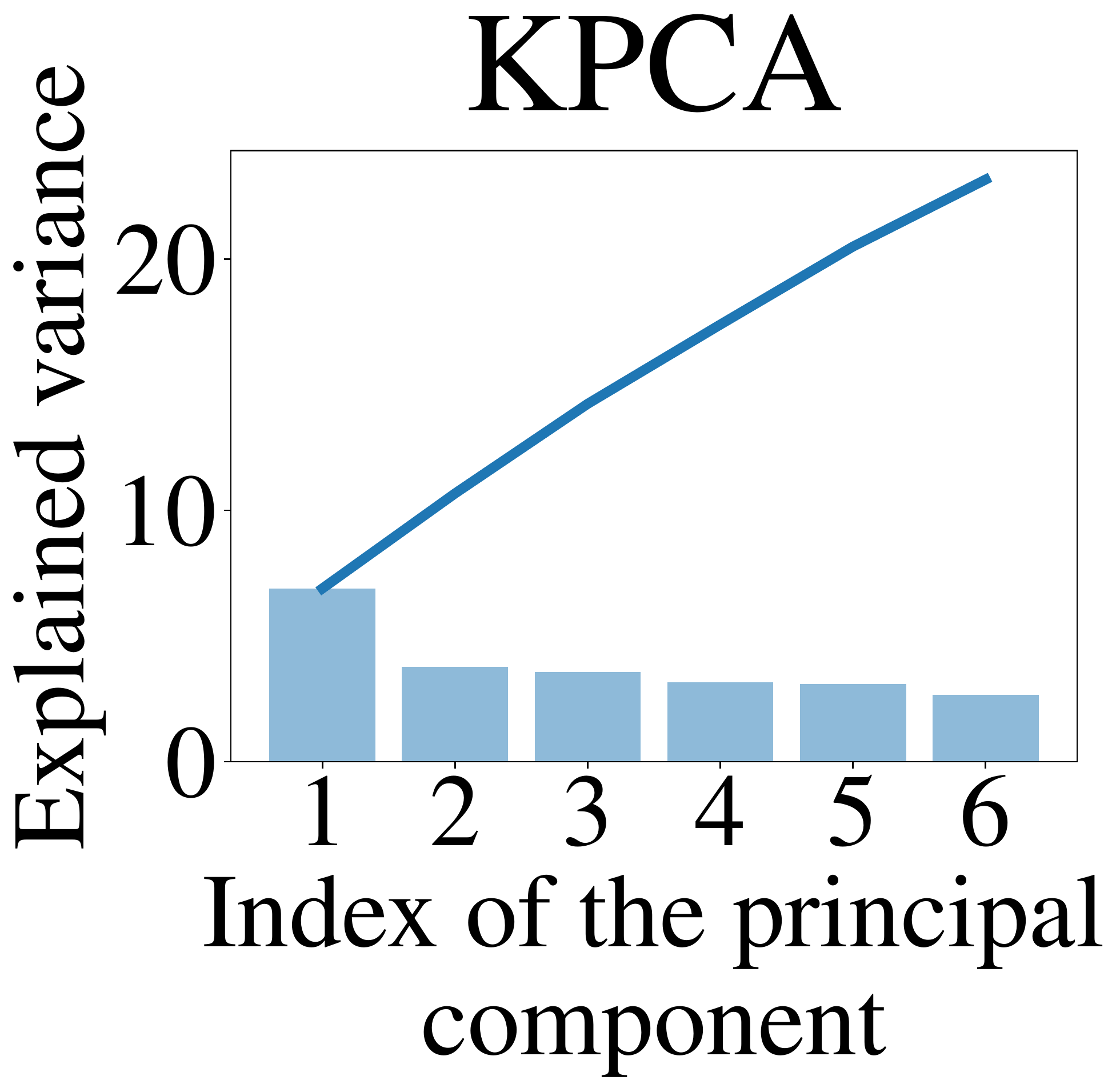}
  \includegraphics[width=0.14\textwidth]{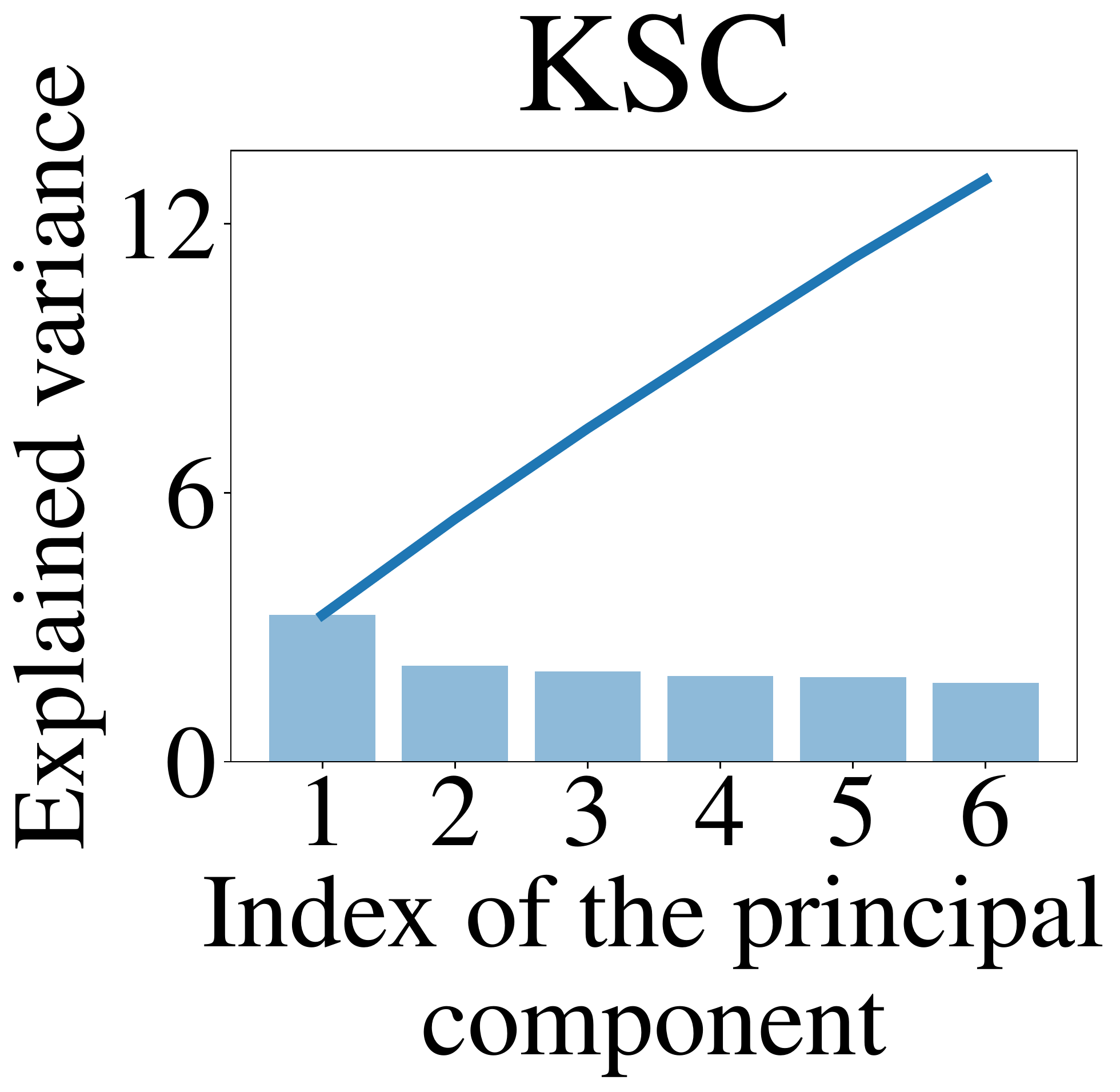}
  \includegraphics[width=0.14\textwidth]{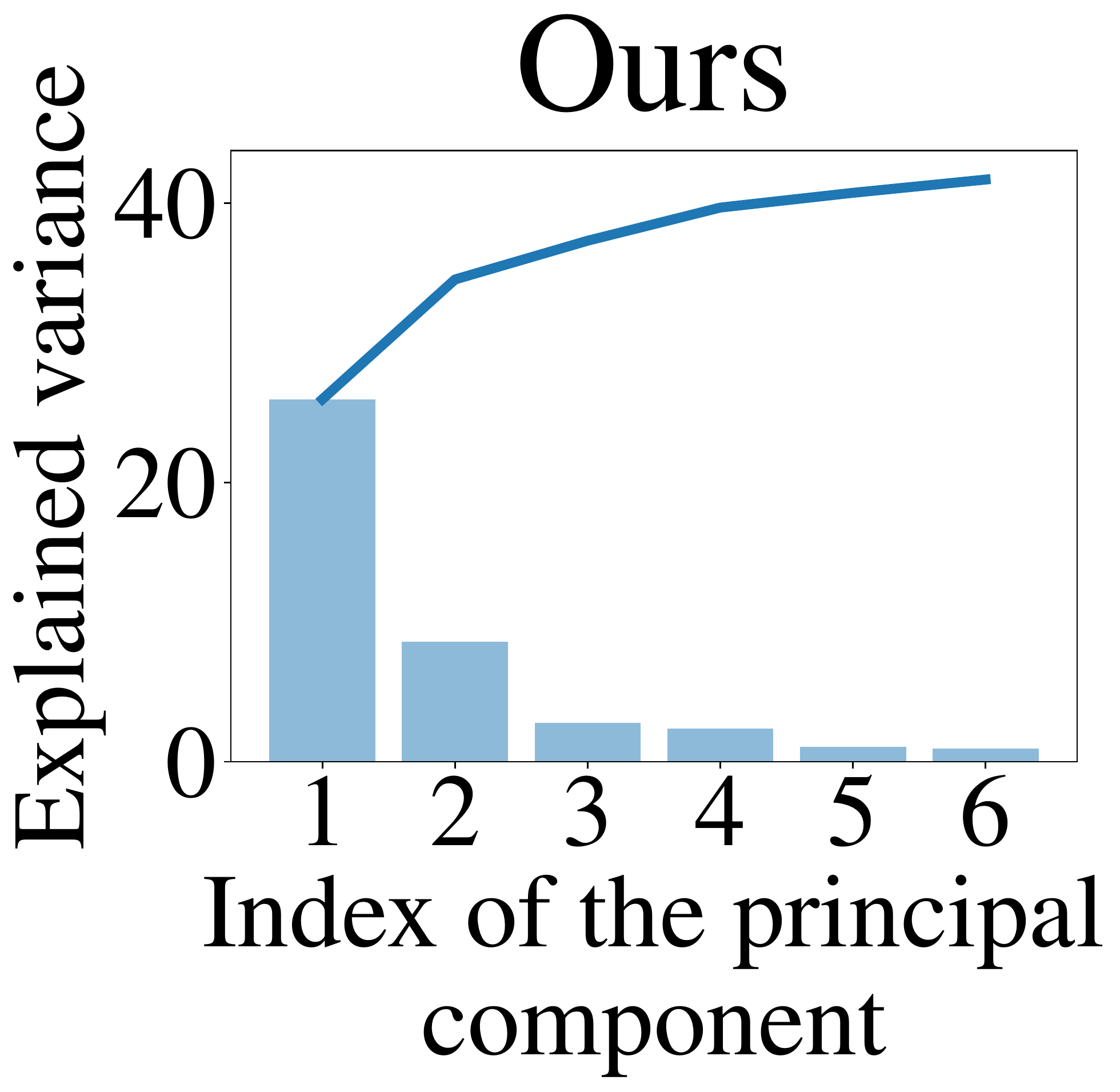}
\caption{Explained variance for Synth 1. Our method captures more information with fewer components and  brings a more informative largest component along with a sharper eigenvalue decay.}
\label{fig:eigs}
\end{figure}

The selected eigenvectors corresponding to the largest eigenvalues in \eqref{eq:dual:second:extension:general} are used for clustering and play a significant role in the performance. We investigate the largest eigenvalues: Figure  \ref{fig:eigs} plots the percentage of explained variance  and the cumulative  variance for Synth 1,  {where KPCA and KSC are applied with attribute concatenation}.

\begin{figure}[ht]
\centering
\begin{subfigure}{.15\textwidth}
  \centering
  \includegraphics[scale=1]{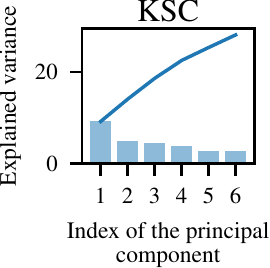}
\end{subfigure}%
\
\begin{subfigure}{.15\textwidth}
  \centering
  \includegraphics[width=\textwidth]{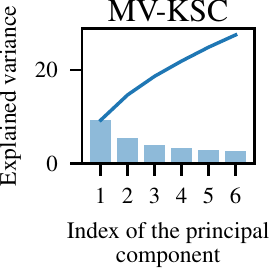}
\end{subfigure}%
\
\begin{subfigure}{.15\textwidth}
  \centering
  \includegraphics[scale=1]{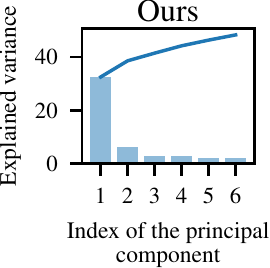}
\end{subfigure}%
\caption{Explained variance for Ads ($k=2$).  This plot also demonstrates our advantages in terms of a sharper eigenvalue decay compared to both the single-view based KSC and the multi-view based MvKSC.}
\label{fig:eigs:ads}
\end{figure}

In Figure \ref{fig:eigs},  the eigenvalue drops distinctively after the first component in all plots, where our method shows the sharpest decay. This dataset has two clusters, so the first eigenvector is sufficient for binary clustering. The percentage of explained variance by the first  component in our method is around 25\%, compared to less than  8\% in KPCA and 5\% in KSC. The advantage of KSC is not distinctive over KPCA here, though KSC employs the degree  matrix to introduce the clustering information. This experiment shows that our method indeed leads to more informative components, ultimately resulting in more separable clusters. For real-world  {data}, an analogous analysis can be performed, where  Ads  is illustrated  and comparisons to MvKSC are also given in Figure \ref{fig:eigs:ads}, showing similar results.

\subsubsection{Latent Space}
\begin{figure}[ht!]
\centering
  \includegraphics[scale=1]{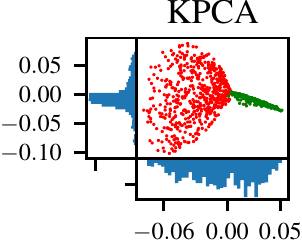}
  \includegraphics[scale=1]{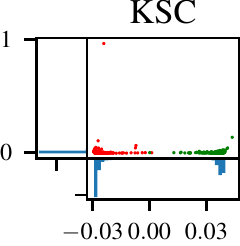}
  \includegraphics[scale=0.77]{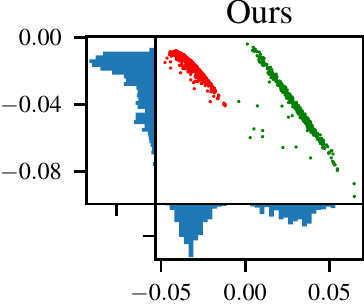}
  \caption{Scatter plot of the latent variable distribution for Synth 1. The histogram of $\bm h^{(1)}$ in KPCA gives an  ungrouped distribution along the horizontal axis, while  the two clusters are well separated in KSC and  our method, and the Gaussian profiles in the histogram of our method are  smoother and more distinctive.}
\label{fig:h}
\end{figure}

\begin{figure}[ht!]
\centering
  \includegraphics[width=0.155\textwidth]{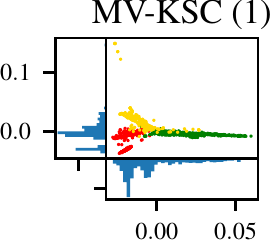}
  \includegraphics[width=0.155\textwidth]{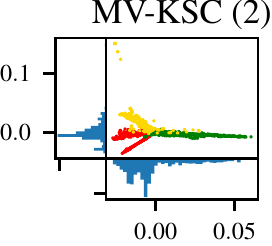}
  \includegraphics[width=0.155\textwidth]{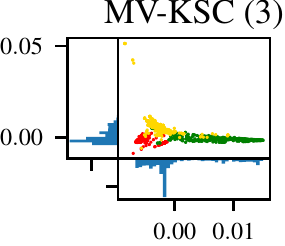}
  \includegraphics[width=0.165\textwidth]{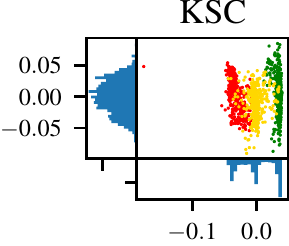}
  \includegraphics[width=0.16\textwidth]{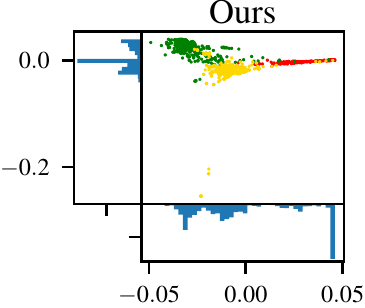}
  \caption{Scatter plot of the latent variable distribution for ImgC, where MvKSC works with  separate latent spaces for  each view. In KSC, the latent variable over the horizontal axis can separate the cluster in green (Aviation) from the those  in yellow (Paintball) and red (Sports), but the other latent variable  does not further identify Paintball and Sports.  Both MvKSC and our method achieve better separations than KSC. Compared to MvKSC, especially on view 3, the clusters in our shared latent space are better separated  and their Gaussian profiles in the histogram are more distinct. }
\label{fig:h:image}
\end{figure}

{With the common latent space, the latent variables, i.e., hidden features, in the constructed model can be directly  visualized herein.}
Figure \ref{fig:h} plots the histogram of the latent variable distributions corresponding to Figure \ref{fig:eigs} on Synth 1. In Figure \ref{fig:h}, the first and the second components are shown in the  horizontal and perpendicular axis, respectively, where the color indicates the ground-truth cluster of each data point. 
Although KPCA and KSC both  show  a similar eigenvalue decay in Figure \ref{fig:eigs}, the  component computed by KSC can formulate better clustering, as shown in Figure \ref{fig:h}. {In KPCA, the first eigenvector shows clustering properties, but its histogram still
shows a single ungrouped distribution rather than two distinct distributions for the clustering as in KSC.} In our method, the  two clusters  in the latent space are well separated and their Gaussian profiles in the histogram are also smoother than KSC, indicating the more informative component  in our method.  This result demonstrates that the proposed common latent space  is effective for multi-view learning in improving cluster separations, and might also facilitate future works, e.g., the generative model by sampling the distributions in the latent space. Further, the real-world dataset ImgC with $k=3$ clusters is 
evaluated in Figure \ref{fig:h:image}, where two eigenvectors are needed for clustering, and thereby the histograms on  both axes matter. Similar to Figure \ref{fig:h},  our method  shows better separations and more distinct  histogram distributions for the clusters on both axes in the latent space. {MvKSC works with a separate latent space for each view, each of which can have varied performances, which is inconvenient for a consistent analysis when dealing with  many views.  In contrast, our method determines a single latent space fusing all views in their dual representations  to reveal the underlying clustering across  views and meanwhile results in more separable clusters, boosting clustering performance and simplyifing pattern discovery.}

\subsubsection{Parameter sensitivity studies} 
\begin{figure}[ht!]
	\centering
		\includegraphics[scale=0.8]{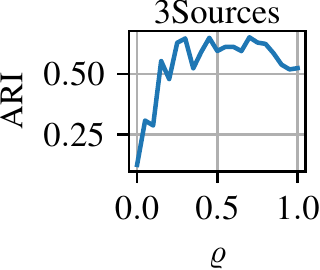}
		\includegraphics[scale=0.8]{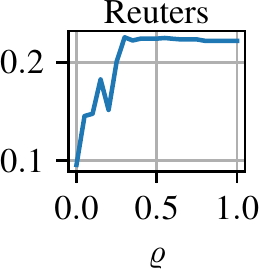}
		\includegraphics[scale=0.8]{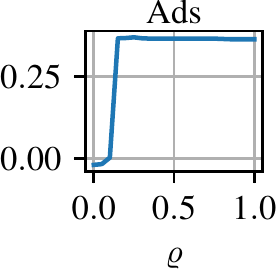}\\
        \vspace{0.1cm}
        \hspace{0.45cm}
		\includegraphics[scale=0.8]{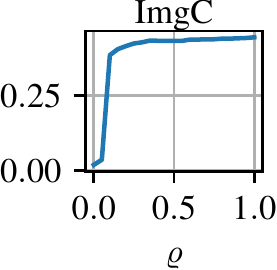}
		\includegraphics[scale=0.8]{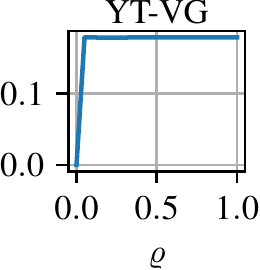}
		\label{fig:rho:game}
		\includegraphics[scale=0.8]{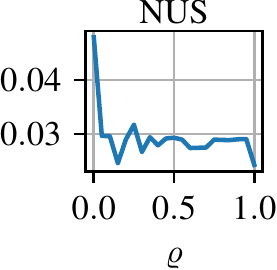}
		\label{fig:rho:nus}
    \caption{{Study on the effects of $\varrho$ in the proposed method}.} 
	\label{fig:varrho}
\end{figure}

\begin{figure}[t!]
\centering
        \includegraphics[width=0.4\columnwidth]{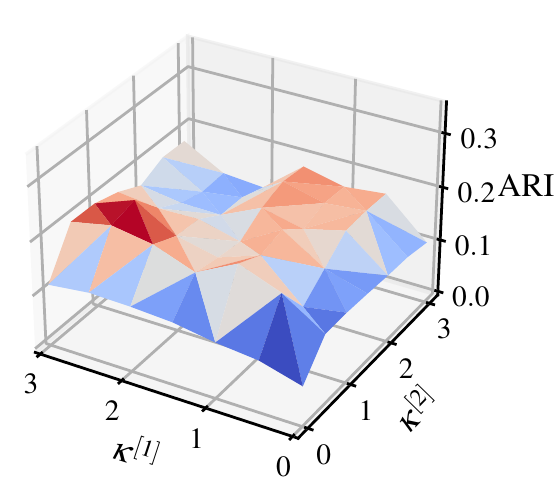}
\caption{{Study on the effects of $\kappa$ in the proposed method}.} 
\label{fig:kappa3d}
\end{figure}

In the proposed  TMvKSCR, 
 $\varrho$ and $\kappa^{[v]}$  can be tuned, as their incorporation  enhances the couplings of views and  integration of fusions.  To better understand the roles of $\varrho$ and $\kappa$, studies are made for these two  hyperparameters  as follows. We firstly evaluate how  $\varrho$,  which combines the fusions {as well as the tensor and matrix representations}, affects the performance for multi-view clustering. In this experiment,  all tested real-world datasets are used for evaluations, with other parameters  fixed  {($\kappa^{[v]}=1$ for $v=1, \dots, V$)}. Figure \ref{fig:varrho} presents the ARI  results with varied $\varrho$. Then, we proceed to study the  $\kappa^{[v]}$, where the Reuters dataset is exemplified for an illustration in Figure \ref{fig:kappa3d}, which plots the  surface of ARI with varying $\kappa^{[v]}, v=1, 2$ and fixing $\kappa^{[3]}=1$ and $\varrho=0.3$.

As shown in Figure \ref{fig:varrho}, very low values of $\varrho$ lead to worse ARI on most datasets, suggesting that  using tensors alone commonly does not result in good clustering.  On the other hand, the optimal $\varrho$ for all datasets is greater than 0, showing the effectiveness of integrating tensor learning for high-order correlations. Hence, a good balance of $\varrho$ can improve the clustering performance. Especially for NUS, the tensor learning can  bring significantly better ARI results, as smaller $\varrho$ gives higher ARI. {$\varrho=0.25$ can be taken as a mild suggestion giving desirable performances for most cases.} For the studies of $\kappa$ in  Figure \ref{fig:kappa3d}, {simply setting  $\kappa^{[1]}=\kappa^{[2]}=1$ already achieves good performance, while the optimal values of $\kappa^{[1]}$ and $\kappa^{[2]}$ are not necessarily exactly 1 and can be adjusted for enhanced performance}, indicating that differently weighting the views can further boost the performance,  which can be particularly meaningful with views varying significantly from each other. In this shown plot, a relatively higher value of $\kappa^{[1]}$ gives better performance. 

\section{Conclusion}\label{sec:conclusion}
In this paper, a novel method is proposed for multi-view spectral clustering  under the framework of RKM. Firstly, {we  introduce a modified weighted conjugate feature duality to formulate a RKM for spectral clustering. Secondly,}  we
propose to utilize shared hidden features with a common latent projection space, which realizes the couplings among different views {and provides a simple alternative to perform interpretation and analysis. Thirdly,}  a tensor-based model  is constructed to simultaneously incorporate feature maps of all views into tensors, which realizes high-order correlations and  generalizes view fusions {without increasing computational complexity in optimization}. {Fourthly, the resulting optimization simply needs to solve a single eigenvalue decomposition problem and the proposed shared latent space results in a computational complexity independent of number of views. Besides,}  the out-of-sample extension of our method enables to infer unseen data without re-training, which also greatly
benefits the efficiency for large-scale cases.  
Numerical experiments on both synthetic and real-world datasets verify the effectiveness regarding clustering accuracy and efficiency of the proposed method. Through further analysis, we show that our method can capture more information in fewer components and that its shared latent space can boost informativeness in terms of data exploration {and visualization}.
In future work, it would be of interest to investigate deep-level RKM  and different tensor techniques  for varied demands.



\section*{Acknowledgments}
This work is jointly supported by ERC Advanced Grant E-DUALITY (787960), KU Leuven Grant CoE PFV/10/002, and Grant  FWO G0A4917N, EU H2020 ICT-48 Network TAILOR (Foundations of Trustworthy AI - Integrating Reasoning, Learning and Optimization), and Leuven.AI Institute. This work was also supported by the Research Foundation Flanders (FWO) research projects G086518N, G086318N, and G0A0920N; Fonds de la Recherche Scientifique — FNRS and the Fonds Wetenschappelijk Onderzoek — Vlaanderen under EOS Project No. 30468160 (SeLMA). 


\ifCLASSOPTIONcaptionsoff
  \newpage
\fi

\bibliographystyle{IEEEtran}
\bibliography{tpami-mvkscrkm}

\clearpage
\newpage

\appendices
\section{More  Experimental Details}\label{appendix:experiment:setup} 

The Synth 1 dataset consists of two-dimensional samples in three views $\{ (\bm x_i^{[1]}, \bm x_i^{[2]}, \bm x_i^{[3]}) \}_{i=1}^{1000}$  generated by a Gaussian mixture model. The cluster means for the first view are [1 1] and [3 4], for the second view are [1 2] and [2 2], and for the third view are [1 1] and [3 3]. The covariance matrices are $\bm \Sigma_1^{[1]}=\begin{bmatrix} 1 & 0.5 \\ 0.5 & 1.5 \end{bmatrix}$, $\bm \Sigma_2^{[1]}=\begin{bmatrix} 0.3 & 0.2 \\ 0.2 & 0.6 \end{bmatrix}$, $\bm \Sigma_1^{[2]}=\begin{bmatrix} 1 & -0.2 \\ -0.2 & 1 \end{bmatrix}$, $\bm \Sigma_2^{[2]}=\begin{bmatrix} 0.6 & 0.1 \\ 0.1 & 0.5 \end{bmatrix}$, $\bm \Sigma_1^{[3]}=\begin{bmatrix} 1.2 & 0.2 \\ 0.2 & 1 \end{bmatrix}$, $\bm \Sigma_2^{[3]}=\begin{bmatrix} 1 & 0.4 \\ 0.4 & 0.7 \end{bmatrix}$. 
The Synth 2 dataset consists of two-dimensional samples in two views $\{ (\bm x_i^{[1]}, \bm x_i^{[2]}) \}_{i=1}^{1000}$. For both views, data are allocated in an imbalanced way, i.e., 80\% of the data points are located quite compactly in the dominant cluster while the remaining data are allocated more loosely in another cluster. More specifically, a point set $(\bm x^{[1]}_i, \bm x^{[2]}_i)$ is generated from a Gaussian mixture model. The cluster means for the first view are [1 1] and [2 2], and for the second view are [2 2] and [1 1]. The covariance matrices are $\bm \Sigma_1^{[1]}=\begin{bmatrix} 0.1 & 0 \\ 0 & 0.3 \end{bmatrix}$, $\bm \Sigma_2^{[1]}=\begin{bmatrix} 1.5 & 0.4 \\ 0.4 & 1.2 \end{bmatrix}$, $\bm \Sigma_1^{[2]}=\begin{bmatrix} 0.3 & 0 \\ 0 & 0.6 \end{bmatrix}$, $\bm \Sigma_2^{[2]}=\begin{bmatrix} 1 & 0.5 \\ 0.5 & 0.9 \end{bmatrix}$. 

For high-dimensional real-world datasets in  Table \ref{tab:real:data:info}, the 3Sources Text dataset consists of news articles from three online sources, corresponding to the three views: BBC, Reuters, and the Guardian \cite{greene2009matrix}.  Reuters \cite{liu2013multi}  is from the  L-Reuters Multilingual dataset \cite{Dua:2019}, which contains documents of six categories written in five different languages.  In Reuters, the English document is taken as the first view, while the corresponding French and German translation are the second view and third view, respectively.  The Ads dataset contains images with hyperlink labeled as advertisement or not advertisement; the first view describes the image, the second view the URL and the last view the anchor URL \cite{kushmerick1999,kolenda2002,Luo2015TensorCC}. The ImgC dataset consists of images of three sports: the first two views represent features of the image and the third view is the caption of the image \cite{kolenda2002}. The YT-VG dataset consists of videos of video games described in three views of text, image and audio features, as described in \cite{madani2012}. The NUS dataset contains images of ten classes described in five views using color histogram features, local self-similarity, pyramid HOG, SIFT, color SIFT and SURF features \cite{Chua2009NUSWIDEAR}.

Regarding  {hyperparameter} selection, we tune the hyperparameters of each tested method by grid search in the ranges suggested by the authors in their papers. In particular, one of the hyperparameters  of COMIC, i.e., the clustering threshold  $\varepsilon$, controls the estimated number of clusters $\hat{k}$. COMIC can normally overestimate the  number of clusters $\hat{k}$ giving the worst performance among all the compared methods. If we keep the estimated number of clusters $\hat{k}$ to the real $k$, the performance is still inferior. Thus, we make a mild balance to give improved results, i.e.,  $\varepsilon$ is tuned between 0 and 1 with $\hat{k}/N \leq 0.1$. The shared hyperparameters in all
methods are tuned under the same settings, e.g., the kernel parameters are tuned in the same range. The RBF kernel  is used for the two synthetic datasets,  Ads, NUS datasets, and the first two views of ImgC datasets. The 3Sources, Reuters, and the third view of ImgC datasets consist of high-dimensional text data, so we employ the normalized polynomial kernel of degree $d$. For the sparse and high-dimensional YT-VG dataset, we employ the linear kernel. The $\sigma^2$ of  RBF kernels is tuned between $e^{-7}$ and $e^7$, the employed degree of the normalized polynomial kernel is $d \in \{1,2\}$, and the $t$ parameter is chosen between $e^{-5}$ and $e^5$. For MvKSC, the assignment can be done by both $\bm e^{[v]}$ and $\bm e_\text{mean}$,  and we report their best results, while in  our method TMvKSCR  we  use the  assignment $\bm e_\text{mean}$  with  $\varrho$ tuned between 0 and 1 and $\kappa^{[v]}$  between 0 and 3. For Co-Reg, the best results of the two schemes, i.e., the
pairwise and  the centroid couplings, are reported. {For SFMC, we use the suggested anchor selection method and the anchor rate of 0.5 as in their paper. 
For MVSCBP, we tune the number of salient points $m$, the $r$ parameter and the number of nearest neighbors.}
The specific hyperparameters owned by
individual methods are tuned separately as described above.

\end{document}